\documentclass{article}

\usepackage{subcaption} % Required for creating subfigures
\usepackage{graphicx}
\usepackage{subcaption}
\usepackage[export]{adjustbox}

\usepackage[preprint]{neurips_data_2023}

\usepackage[utf8]{inputenc} % allow utf-8 input
\usepackage[T1]{fontenc}    % use 8-bit T1 fonts
\usepackage[hidelinks]{hyperref}       % hyperlinks
\usepackage{url}            % simple URL typesetting
\usepackage{booktabs}       % professional-quality tables
\usepackage{amsfonts}       % blackboard math symbols
\usepackage{nicefrac}       % compact symbols for 1/2, etc.
\usepackage{microtype}      % microtypography
\usepackage{xcolor}         % colors
\usepackage{siunitx}
\usepackage{algorithm}
\usepackage{algpseudocode}
\usepackage{float}
\usepackage{hyperref}

\usepackage[capitalize]{cleveref}

\usepackage{scalerel,xparse}
\NewDocumentCommand\emojiwineglass{}{
    \scalerel*{
        \includegraphics{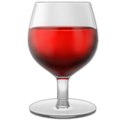}
    }{X}
}
\NewDocumentCommand\emojiwinebottle{}{
    \scalerel*{
        \includegraphics{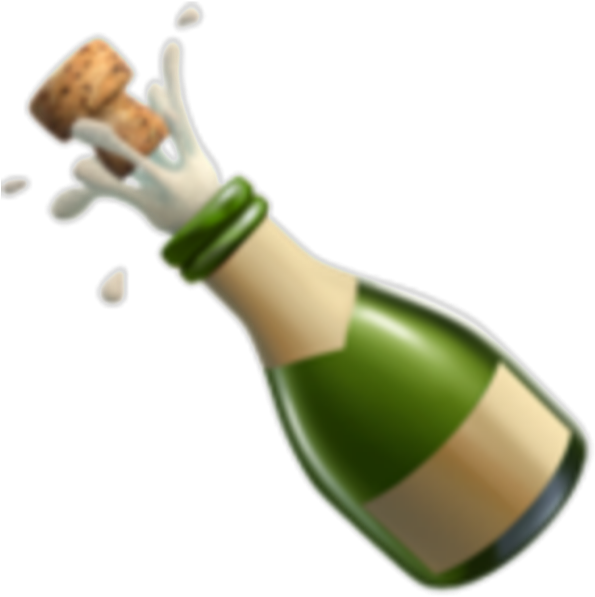}
    }{X}
}
\NewDocumentCommand\emojipurliquid{}{
    \scalerel*{
        \includegraphics{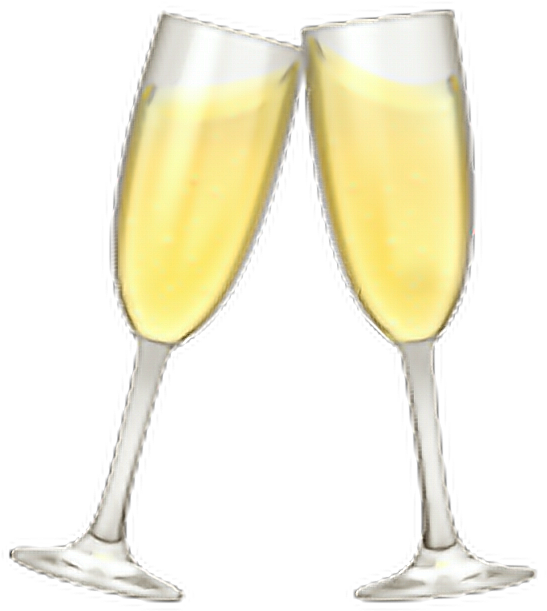}
    }{X}
}

\NewDocumentCommand\emojichampageglass{}{
    \scalerel*{
        \includegraphics{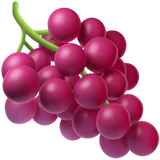}
    }{X}
}

\title{Learning to Taste\emojiwineglass: A Multimodal Wine Dataset}

\author{%
  Thoranna Bender\textsuperscript{\emojichampageglass}  \ \ Simon Moe Sørensen\textsuperscript{\emojichampageglass} \ \ Alireza Kashani\emojiwineglass \ \
  K. Eldjarn Hjorleifsson\textsuperscript{\emojiwinebottle} \\ \textbf{Grethe Hyldig\textsuperscript{\emojichampageglass} \ \ Søren Hauberg\textsuperscript{\emojichampageglass} \ \ Serge Belongie \textsuperscript{\emojipurliquid} \  Frederik Warburg\emojichampageglass} \\ 
  \\
  \textsuperscript{\emojichampageglass} Technical University of Denmark  \textsuperscript{\emojiwineglass} Vivino  \\ \textsuperscript{\emojiwinebottle} California Institute of Technology \textsuperscript{\emojipurliquid} University of Copenhagen \\
}

\begin{document}

\maketitle

\begin{abstract}
  % The abstract paragraph should be indented \nicefrac{1}{2}~inch (3~picas) on
  % both the left- and right-hand margins. Use 10~point type, with a vertical
  % spacing (leading) of 11~points.  The word \textbf{Abstract} must be centered,
  % bold, and in point size 12. Two line spaces precede the abstract. The abstract
  % must be limited to one paragraph.
  We present \textit{WineSensed}, a large multimodal wine dataset for studying the relations between visual perception, language, and flavor. The dataset encompasses 897k images of wine labels and 824k reviews of wines curated from the Vivino platform. It has over 350k unique bottlings, annotated with year, region, rating, alcohol percentage, price, and grape composition. We obtained fine-grained flavor annotations on a subset by conducting a wine-tasting experiment with 256 participants who were asked to rank wines based on their similarity in flavor, resulting in more than 5k pairwise flavor distances. We propose a low-dimensional concept embedding algorithm that combines human experience with automatic machine similarity kernels. We demonstrate that this shared concept embedding space improves upon separate embedding spaces for coarse flavor classification (alcohol percentage, country, grape, price, rating) and aligns with the intricate human perception of flavor.\looseness=-1
  
  %labelThe data is collected from the Vivino platform and via 7 human-centered The dataset encompasses, e that merges human annotations of perceived similarities between 110 distinct wines, collected from 256 individuals, with text embeddings derived from wine reviews. Using the Napping technique, participants tasted five wine samples and arranged them on an A3 sheet of paper according to their flavor similarities. These spatial representations were subsequently quantified on a continuous scale to create similarity annotations. By combining human annotations with text embeddings, WineSensed enables the development of a joint embedding space that improves wine classification performance across various attributes, such as country of origin, alcohol percentage, price, and rating. Our experiments demonstrate the effectiveness of this joint embedding in enhancing classification tasks for wines. Specifically, a classifier's performance on the joint embedding space to predict the country of origin is 22.7\% better than when using just reviews and 107.7\% better than when using only human-annotations. The WineSensed dataset is publicly available to foster further research and applications in wine classification, recommendation, and beyond. We hope that our dataset will inspire the development of new methods and models that capitalize on both human perception and textual information to enrich the understanding and modeling of sensory experiences across various domains.
\end{abstract}

\section{Introduction}
Vision, language, audio, touch, smell, and taste are sensory inputs that ground humans in a shared representation, which enables us to interact, converse, and create. Recent advances in multimodal learning have shown that combining diverse modalities in a shared representation leads to useful and better-grounded models~\citep{girdhar2023imagebind, chen2023sound}. Inspired by recent progress, we propose to add flavor to the list of modalities used to learn shared representations. 

As a first step towards modeling flavor, we focus on wine since (1) wines have been studied for centuries, (2) their flavors have been carefully categorized, and (3) classification systems exist to ensure that flavor is near-consistent across bottles of the same unique bottling. 

% We propose a multimodal wine dataset, WineSensed, containing images, user reviews, and fine-grained flavor annotations. We annotate flavor in wine through a large sensory study, where more than 250 participants were asked to judge their perceived taste similarities of wines. To conduct this study, we are inspired by the ``Napping'' methodology~\citep{pages2005collection} commonly used in food science to conduct consumer surveys~\citep{kim2013cross, ribeiro2020optimising}. \cref{fig:contribution}(right) illustrates how participants were asked to place wines on a sheet of paper based on how similar they perceive the taste of the wines. We combine this annotated fine-grained flavor data with images of wine labels, user reviews, and meta-data such as origin, alcohol percentage, price, and grape composition into a large multimodal data. The image, user reviews, and meta-data are curated from the Vivino platform, an online social media for wine enthusiasts.

%In the food sciences, there is a growing need for comprehensive and nuanced datasets. To address this,
We bridge the gap between the machine learning and food science communities by presenting WineSensed, a multimodal wine dataset that consists of images, user reviews, and flavor annotations. Our motivation is twofold. On one hand, internet photos and user reviews are a scalable source of data, offering abundant, diverse, and easily accessible insights into wine qualities. On the other hand, human flavor annotations, while not as scalable, provide a more direct and granular understanding of the wines' flavor profile. By combining these resources, we aim to capture the best of both worlds, yielding a richer, more intricate dataset.% \serge{nice wording here}

\begin{figure}
    \centering
    \includegraphics[width=\textwidth]{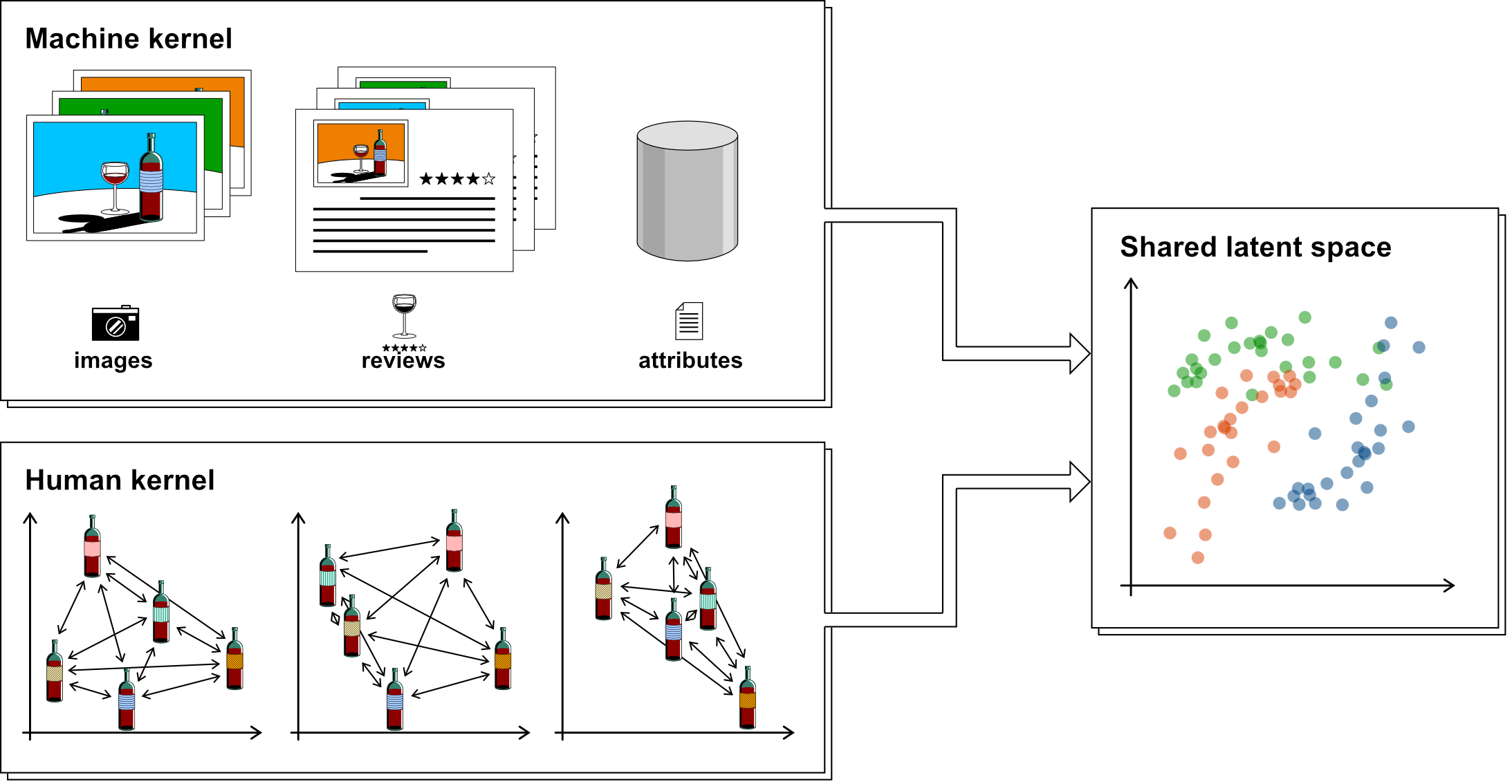}
    \caption{\textbf{Flavor as an additional data modality.} The WineSensed dataset consists of a large collection of images, user reviews, and metadata about unique bottlings (upper left). In a large user study, we collected flavor annotations of over 100 wines using the ``Napping'' method~\citep{pages2005collection}, where participants were asked to place wines on a sheet of paper based on their perceived taste similarity (lower left). We propose an algorithm to combine these data modalities into a shared representation (right) and find that using taste annotations as an additional modality improves performance in downstream tasks. %\serge{excellent caption}
    }
    \label{fig:contribution}
\end{figure}

We organized a large sensory study to obtain human-annotated flavor profiles of the wines. The study applies the ``Napping'' methodology~\citep{pages2005collection}, which is commonly used to conduct consumer surveys~\citep{kim2013cross, ribeiro2020optimising}. In this study, 256 participants annotated their perceived taste similarities of various wines. In \cref{fig:contribution}, the ``human kernel'' illustrates how participants were instructed to place wines on a sheet of paper based on how similar they perceived their flavor to be. The Napping method enabled us to annotate wine flavors with a high level of detail and harness the perception of a broad spectrum of individuals. It scales well, as asking a participant to annotate five wines yields 10 pairwise annotations. All participants combined annotated more than 5k flavor distances.

%Per particpant with uptained 30 pairwise distances (from ask

%Furthermore, by asking participants to consider the similarities of five wines at a time, we obtained up to 30 pairwise annotations per participant, resulting in more than 5k pairwise flavor distances.  \fred{not super happy with this last sentence.}

%Furthermore, the method scales well. It yields 30 pairwise annotations per participant by asking each participant to consider five wines at a time
%The method scales well, yielding 30 pairwise annotations per participant and over 5,000 flavor distances, by asking participants to consider the similarities of five wines at a time.

To complement these annotations, we curate images of wine labels, user reviews, and wine attributes (country of origin, alcohol percentage, price, and grape composition) from the Vivino platform, a popular online social network for wine enthusiasts.\footnote{\url{https://vivino.com}} WineSensed, therefore, represents a large, multimodal dataset that merges user-generated content with sensory assessments, bridging the gap between subjective consumer perception and objective flavor profiles.

Along with the dataset, we propose \emph{Flavor Embeddings from Annotated Similarity \& Text-Image} (FEAST) that leverages recent developments in large multimodal models to embed user reviews and images of wine labels into a low-dimensional, latent representation that contains semantic and structural information that correlates with taste. Our model aligns this representation with the flavor annotations from our user study. We find that this combined representation yields a ``flavor space'' that models coarse flavor concepts like alcohol percentage, country, grape, and the year of production, while also being aligned with more intricate human perception of flavor. 

Experimentally, we find (1) that using the pairwise distances (rather than ordering) of the annotated wines improves the flavor representation, which confirms the established methodology in food science, and validates our annotation process. (2) We discover that using multiple data modalities (images, text, and flavor annotations) boosts the flavor representations, highlighting the usefulness of our multimodal dataset. (3) Finally, we show that the proposed multimodal model produces a flavor space with a high alignment with humans' perception of flavor. %\fred{not happy with this section. remember to revisit.}

\section{Background and related work}

\textbf{Multimodal representations.} Learning a shared representation between modalities can reveal useful representations that generalize well and appear grounded in reality. Pioneering work~\citep{de1994learning} proposes to learn the correlation between vision and audio. A number of deep learning methods propose to use large collections of weakly annotated data to learn shared vision-language representations~\citep{joulin2016learning, desai2021virtex, radford2021learning, mahajan2018exploring}, shared audio-text representations~\citep{agostinelli2023musiclm}, shared vision-audio representations~\citep{ngiam2011multimodal, owens2016visually, arandjelovic2017look, narasimhan2022strumming, hu2022mix}, shared vision-touch~\citep{yang2022touch} representations, or shared sound and Inertial Measurement Unit (IMU) representations~\citep{chen2023sound}. Recently, ImageBind~\citep{girdhar2023imagebind} showed that images can bind multiple modalities (images, text, audio, depth, thermal, IMU) into a shared representation. While recent advances in other areas of multimodal learning have been fueled by large datasets, the difficulty of quantifying and collecting high-quality flavor data has made it challenging for the machine learning community to develop similar representations for flavor.

% research on flavour in food sciences

\textbf{Quantifying flavor.} Understanding and engineering \emph{flavor} is a central part of food science and essential in the quest towards healthy and sustainable food production \citep{savage2012technology}, but the use of machine learning methods to this end is still in its infancy. \citet{fuentes2019modeling} found a correlation between seasonal weather characteristics, and wine quality and aroma profiles, thereby verifying what wine producers have long held to be true. Similarly, \citet{gupta2018selection} found that sulfur dioxide, pH, and alcohol levels are useful for predicting wine quality. Due to the difficulty of gathering quality perception data, much work focuses on how `low-level' chemical aspects related to `high-level' taste properties, e.g.\@ in assessing the quality of chocolate and beer \citep{gunaratne2019chocolate, gonzalez2018assessment}. 

Analyzing a person's perception of wine is challenging due to the complex nature of flavor, which remains ill-understood, and the difficulty in obtaining consistent verbal descriptions of taste across individuals. Napping~\citep{pages2005collection} is the \textit{de facto} method to analyze perceived taste in consumer surveys. Participants receive taste samples and are instructed to place them on a sheet of paper based on how similar they perceive their taste to be, with closer meaning more similar. Such experiments are usually conducted with 10-25 participants and less than 20 variants of a product~\citep{giacalone2013consumer, pages2010sorted, mayhew2016napping}. In this study, we scale this data collection process to 256 participants and 108 unique bottlings of red wine, resulting in over 400 napping papers collected and more than 5k annotated flavor distances. In contrast to previous works~\citep{giacalone2013consumer, pages2010sorted, mayhew2016napping} our objective is to incorporate taste as one of the modalities that contribute to the shared representations for improved grounding of machine learning models.

\begin{figure}
    \centering
    \includegraphics[width=\textwidth]{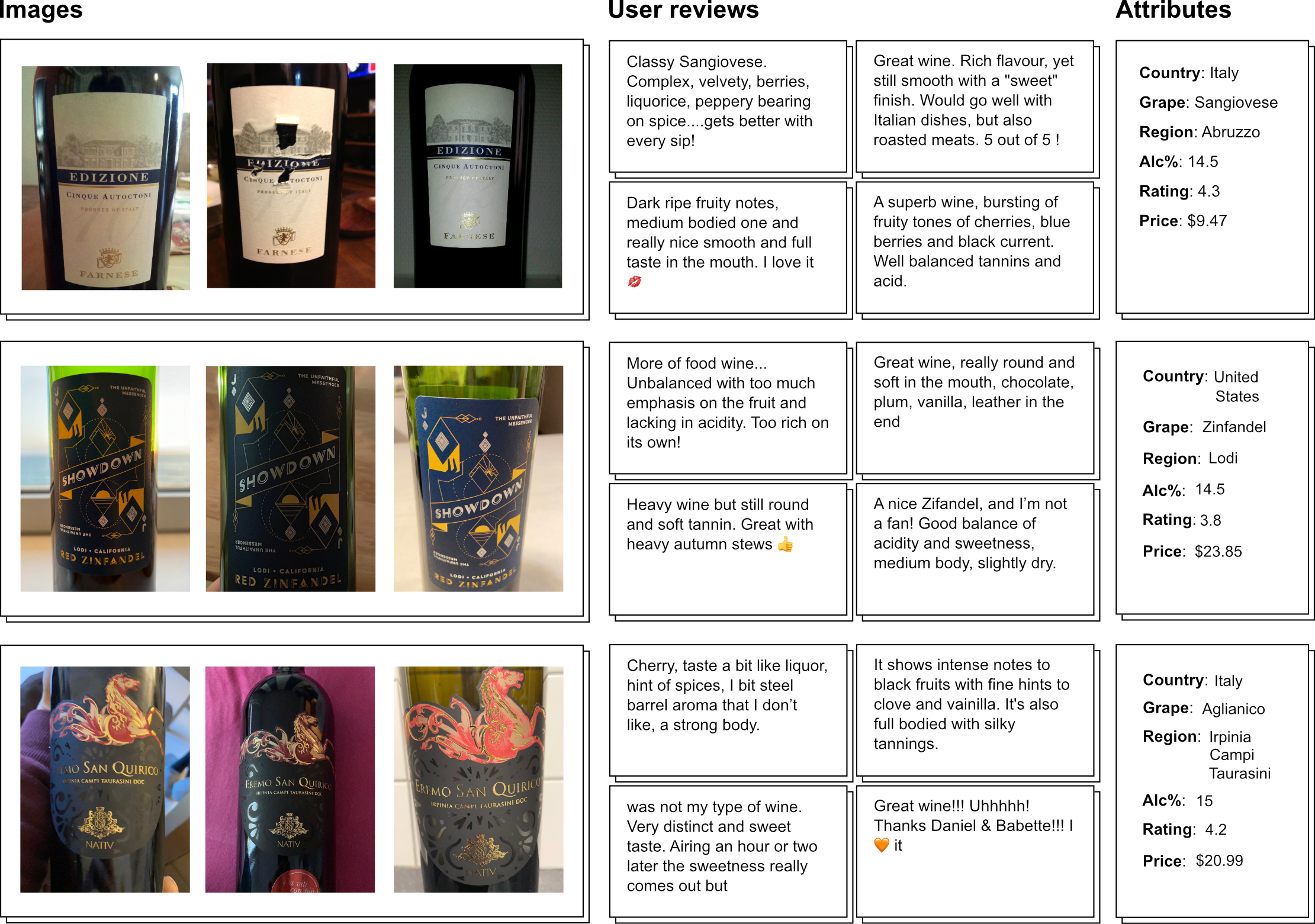}
    \caption{\textbf{Examples from WineSensed.} The dataset consists of images of wine labels, user-generated reviews, per-wine attributes (country, grape, region, alcohol percentage, rating, price), and flavor annotations. Here are examples of the images, reviews, and attributes. }
    \vspace{-0.7cm}
    \label{fig:datasetoverview}
\end{figure}

% napping, snack
\textbf{Human kernel learning.} Annotating flavor with Napping~\citep{pages2005collection} does not provide image-flavor or text-flavor correspondences but rather relative flavor similarities between sampled products. According to~\citep{miller2019explanation} humans are better at describing abstract concepts such as taste with contrastive questions, such as \textit{``does wine X taste more similar to wine Y or Z?''} For this reason, the machine learning community has used contrastive questions in multiple settings, e.g., for understanding how humans perceive light reflection from surfaces by presenting annotators with image triplets depicting the Stanford Bunny with varying material properties~\citep{agarwal2007generalized}, to produce a genre embedding of musical artists~\citep{van2012stochastic}, and for discovering underlying narratives in online discussions~\citep{christensen2022searching}. Most relevant to our work is SNaCK~\citep{wilber2015learning}, which presents annotators with image triplets depicting foods and asked which two of them taste more similar, to obtain flavor triplets. They proposed to combine this high-level human flavor understanding with low-level image statistics to learn food concepts, e.g., that even though guacamole and wasabi look similar, their taste is not. Having humans annotate image triplets of foods works well for coarse concepts, but does not encompass nuanced differences in taste. In this work, we focus on the much finer-grained taste difference found in wines. These nuances and the complex nature of wine tasting, which involves taste \emph{and} smell, are not easily conveyed through text or images.

%Reviews are subjective and vary based on personal factors and context, leading to inconsistent flavour profiles. Moreover, they might not always offer detailed flavour descriptions, focusing instead on aspects like preference, occasion, etc. Similarly, while images of wine labels provide some contextual clues, they fall short of capturing the intricate combination of tastes and aromas in each wine.

\textbf{Flavor datasets.} The machine learning community has produced numerous food datasets for classifying which meal is in an image~\citep{bossard2014food, min2020isia}, retrieving a recipe given an image~\citep{salvador2017learning, li2022mealrec}, or predicting the origin of wines~\citep{Dua:2019}. While it is possible to extract coarse information about taste from such datasets~\citep{wilber2015learning}, they do not encompass higher resolution details of taste, such as the differences between a Cabernet Sauvignon and Pinot Noir. 

Similarly, the food science community has developed many datasets for understanding and predicting food flavors, nutrient content, and chemistry. Flavornet~\citep{arn1998flavornet}, a dataset on human-perceived aroma compounds, explores partly how smells relate to perceived bitterness or fruitiness in a wine. However, its limitation is its lack of context linking these odors to specific wine varieties and its limited focus on flavor aspects. FoodDB~\citep{harrington2019nutrient} offers comprehensive information on a wide variety of food, its nutrient contents, potential health effects, and macro and micro constituents. However, it lacks user-generated reviews and sensory data, which are crucial for understanding the subjective human perception of food and wine. The Wine Data Set \citep{Dua:2019} focuses on wines, but only contains wines originating from one region in Italy, limiting the dataset's ability to capture the broader diversity of flavor profiles of wines from various regions worldwide. Furthermore, \citet{Dua:2019} solely incorporate the chemical compounds present in each wine, without annotations of flavors and information associating specific wines with each chemical compound. In contrast to previous work, we present a multimodal dataset that contains a large corpus of images and reviews, as well as human-annotated flavor similarities.

%\fred{I would be good that the last sentence talks about our work and how it is unique., e.g. In contrast, we present a multimodal dataset than contains 10x more flavor annotations than the previous largest dataset...}

% Experiencing flavor is a complicated multisensory experience. Flavors are intricately linked to other senses such as smell, texture and sight \cite{TODO: cite it}. Moreover, the perception of flavor is highly individual, influenced by personal preferences, genetic factors, cultural background, and previous experiences \cite{TODO: cite it}. While reviews and images can provide valuable insights into the various aspects of flavor, human annotations on flavor similarity offer a level of depth and nuance that is non existent in the other modalities. Reviews are often influenced by factors beyond taste, such as service quality, ambiance, or personal biases, making comparisons difficult. Images, on the other hand, can only provide visual information, which is just one aspect of the multisensory experience of flavor.

\section{The \textit{WineSensed} dataset} \label{sec:dataset}

We present WineSensed, a large, multimodal wine dataset that combines human flavor annotations, images, and reviews. In this section, we provide an overview of the curation process for each of these modalities.

  \begin{figure}
      \centering
      \begin{subfigure}{0.13\textwidth}
        \centering
        \includegraphics[width=\linewidth]{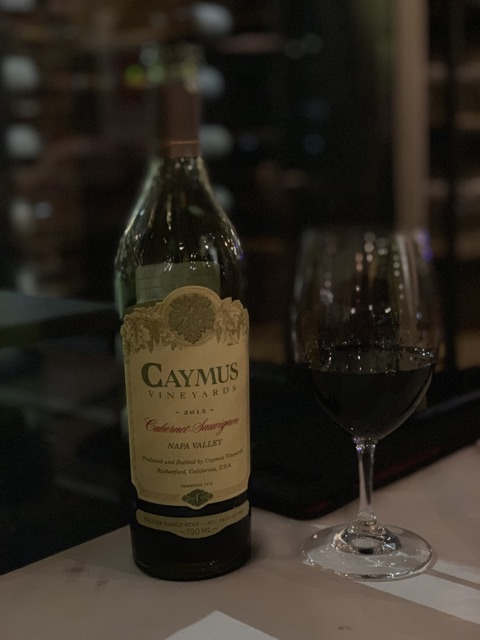}
      \end{subfigure}
      \begin{subfigure}{0.13\textwidth}
        \centering
        \includegraphics[width=\linewidth]{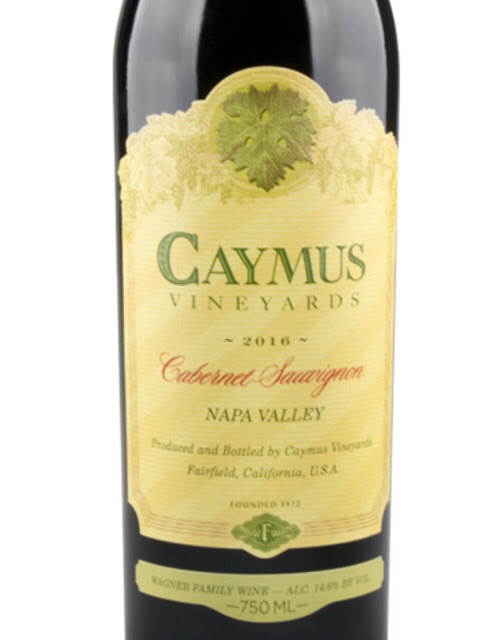}
      \end{subfigure}
      \begin{subfigure}{0.13\textwidth}
        \centering
        \includegraphics[width=\linewidth]{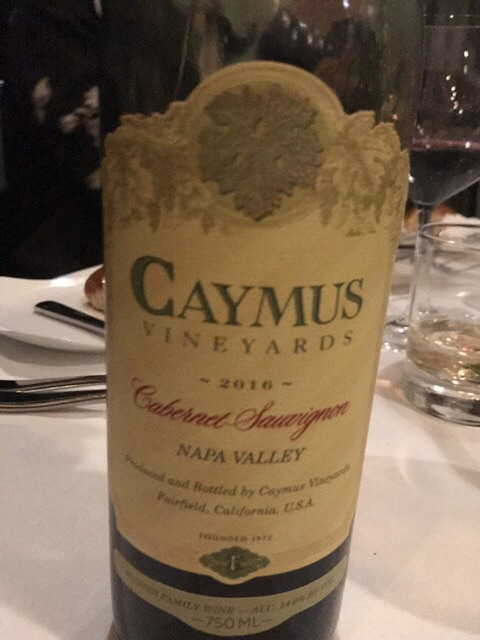}
      \end{subfigure}
      \begin{subfigure}{0.13\textwidth}
        \centering
        \includegraphics[width=\linewidth]{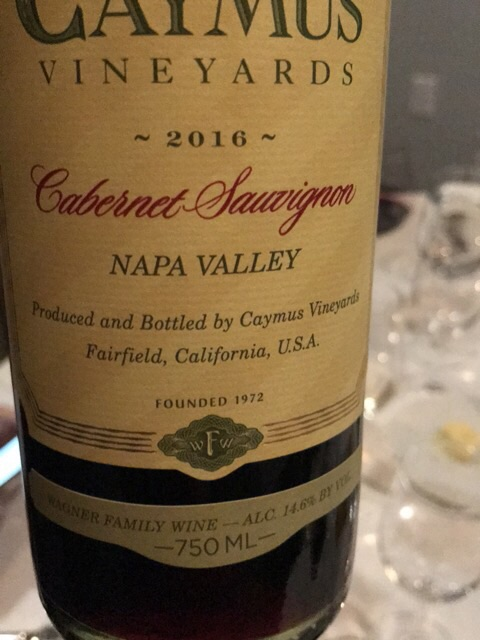}
      \end{subfigure}
      \begin{subfigure}{0.13\textwidth}
        \centering
        \includegraphics[width=\linewidth]{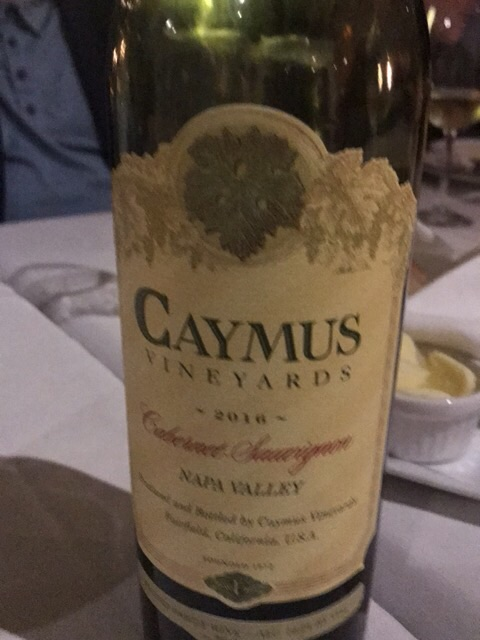}
      \end{subfigure}
      \begin{subfigure}{0.13\textwidth}
        \centering
        \includegraphics[width=\linewidth]{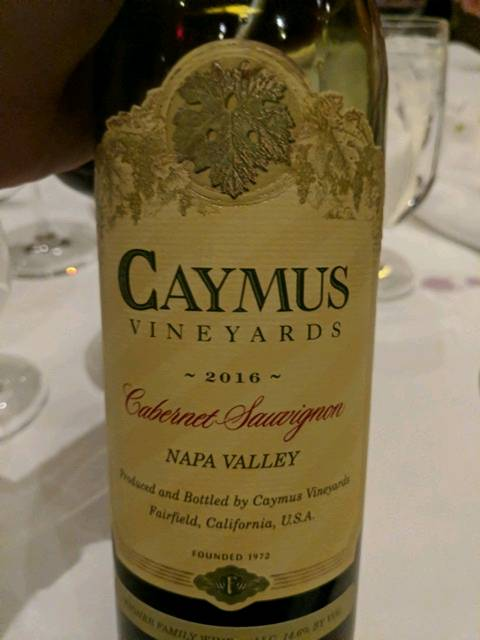}
      \end{subfigure}
      % \begin{subfigure}{0.13\textwidth}
      %   \centering
      %   \includegraphics[width=\linewidth]{images/vintage_samples/vintage_sample-006.png}
      % \end{subfigure}
      \begin{subfigure}{0.13\textwidth}
        \centering
        \includegraphics[width=\linewidth]{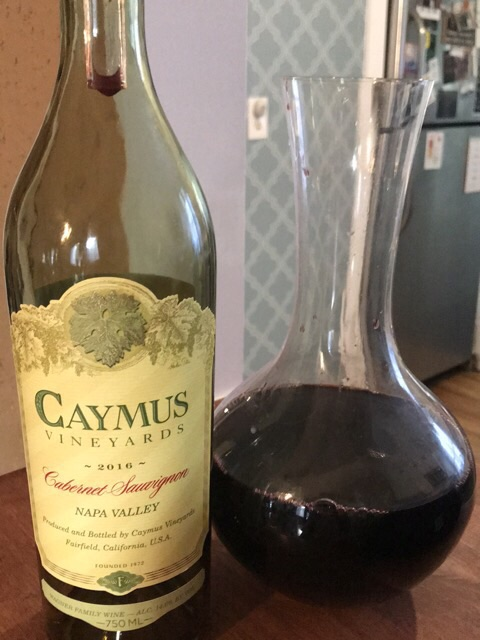}
      \end{subfigure} \vspace{0.1cm}
      \centering
      \begin{subfigure}{0.13\textwidth}
        \centering
        \includegraphics[width=\linewidth]{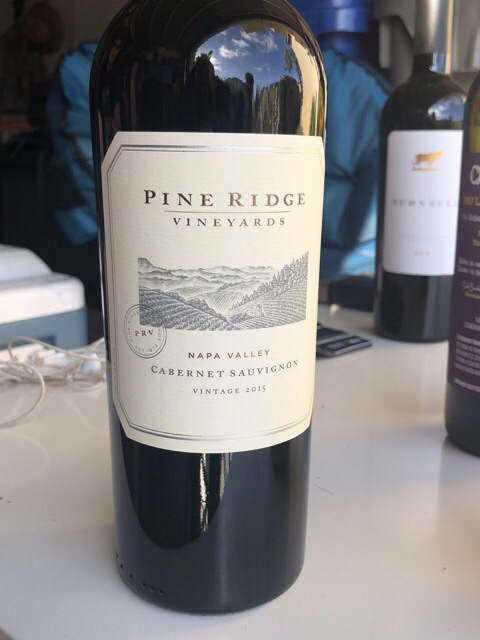}
      \end{subfigure}
      \begin{subfigure}{0.13\textwidth}
        \centering
        \includegraphics[width=\linewidth]{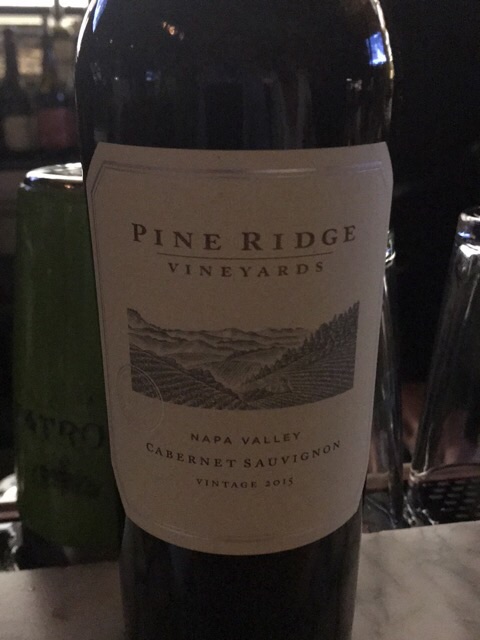}
      \end{subfigure}
      \begin{subfigure}{0.13\textwidth}
        \centering
        \includegraphics[width=\linewidth]{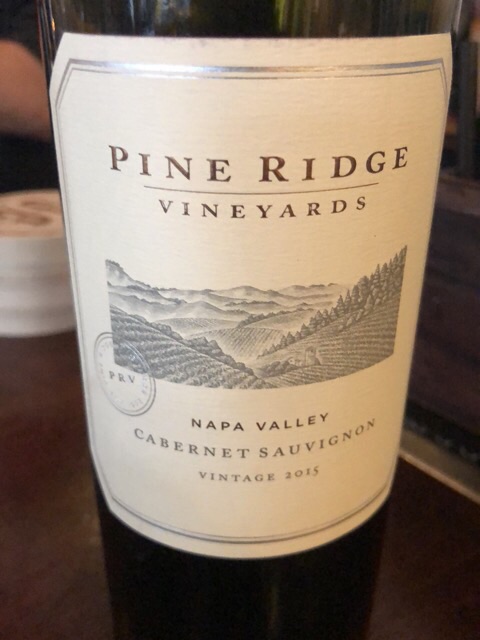}
      \end{subfigure}
      \begin{subfigure}{0.13\textwidth}
        \centering
        \includegraphics[width=\linewidth]{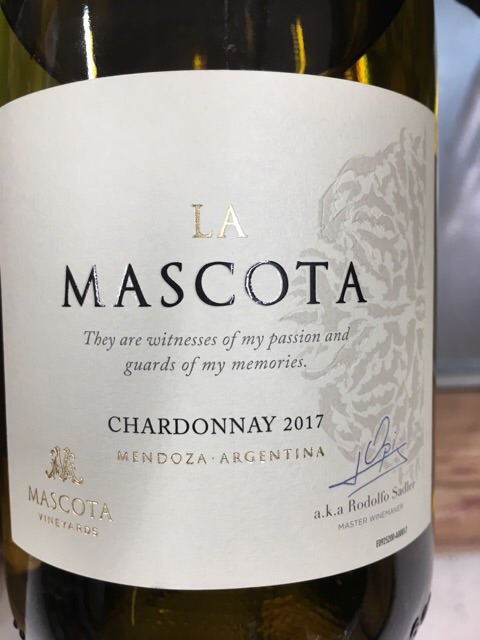}
      \end{subfigure}
      \begin{subfigure}{0.13\textwidth}
        \centering
        \includegraphics[width=\linewidth]{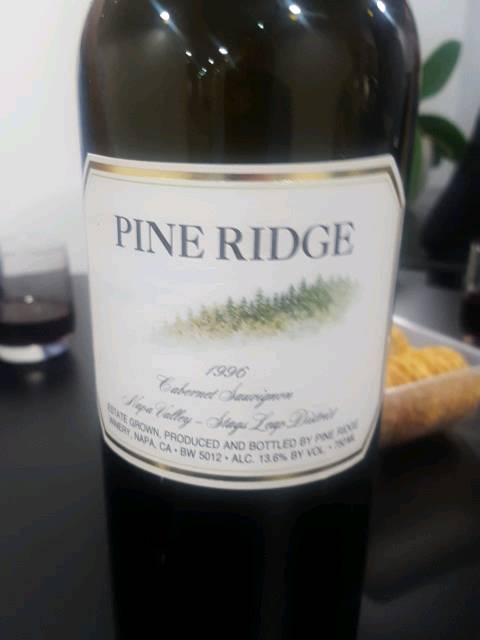}
      \end{subfigure}
      \begin{subfigure}{0.13\textwidth}
        \centering
        \includegraphics[width=\linewidth]{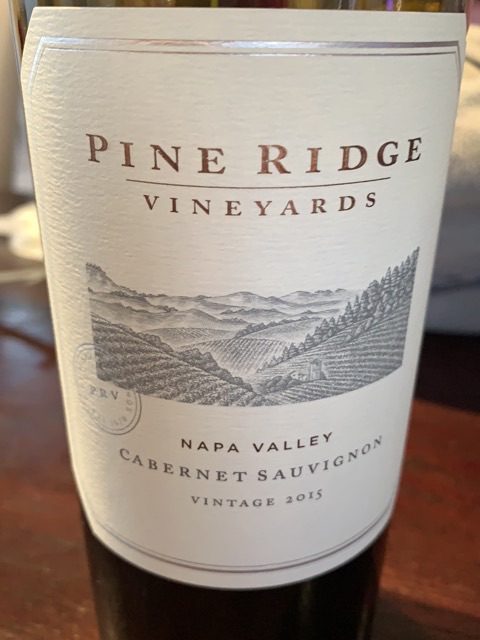}
      \end{subfigure}
      \begin{subfigure}{0.13\textwidth}
        \centering
        \includegraphics[width=\linewidth]{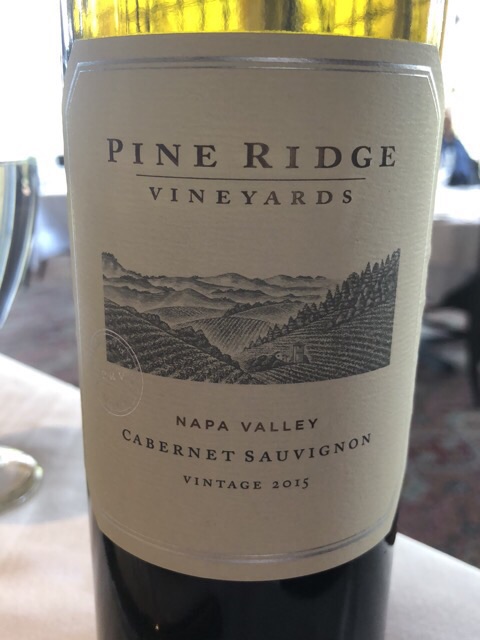}
      \end{subfigure} \vspace{0.1cm}
      \begin{subfigure}{0.13\textwidth}
        \centering
        \includegraphics[width=\linewidth]{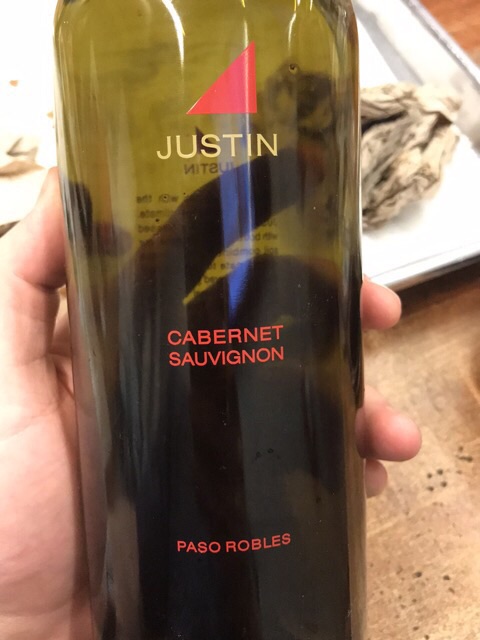}
      \end{subfigure}
      \begin{subfigure}{0.13\textwidth}
        \centering
        \includegraphics[width=\linewidth]{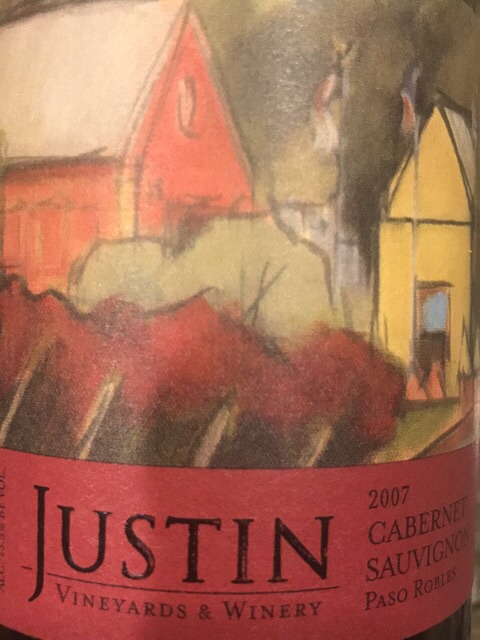}
      \end{subfigure}
      \begin{subfigure}{0.13\textwidth}
        \centering
        \includegraphics[width=\linewidth]{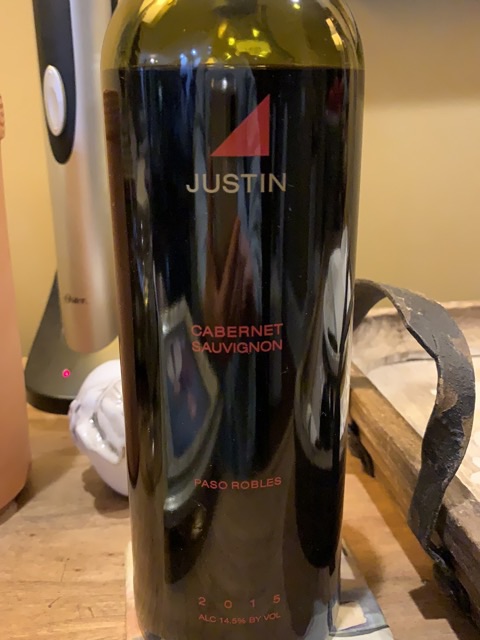}
      \end{subfigure}
      \begin{subfigure}{0.13\textwidth}
        \centering
        \includegraphics[width=\linewidth]{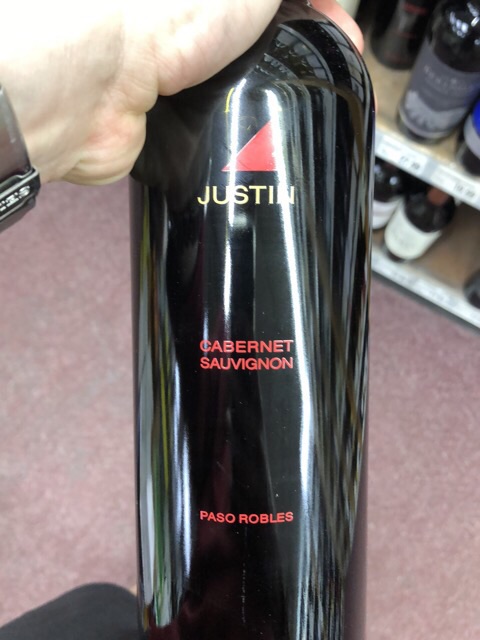}
      \end{subfigure}
      \begin{subfigure}{0.13\textwidth}
        \centering
        \includegraphics[width=\linewidth]{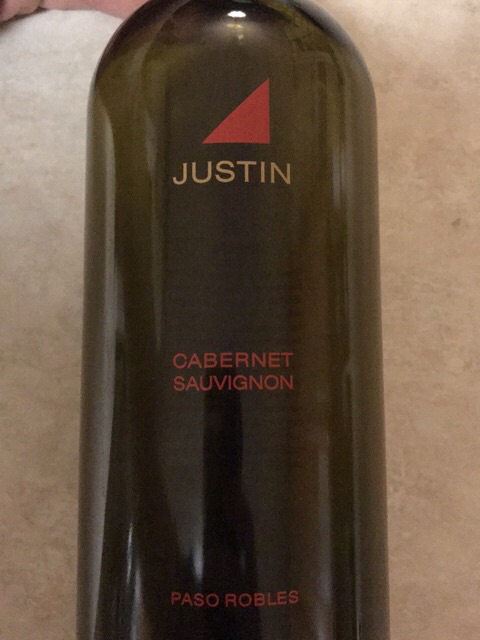}
      \end{subfigure}
      \begin{subfigure}{0.13\textwidth}
        \centering
        \includegraphics[width=\linewidth]{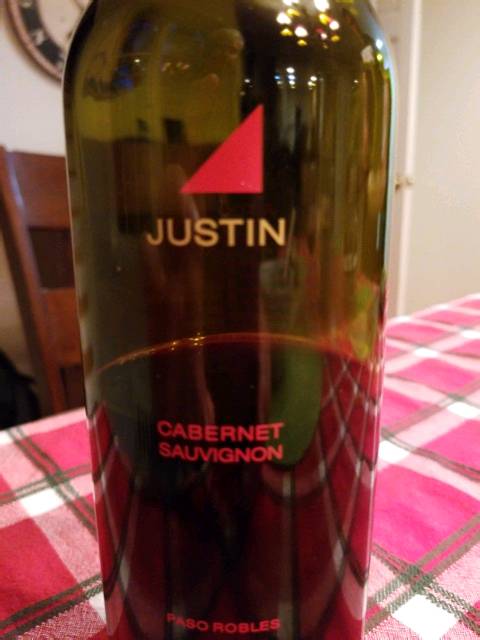}
      \end{subfigure}
      \begin{subfigure}{0.13\textwidth}
        \centering
        \includegraphics[width=\linewidth]{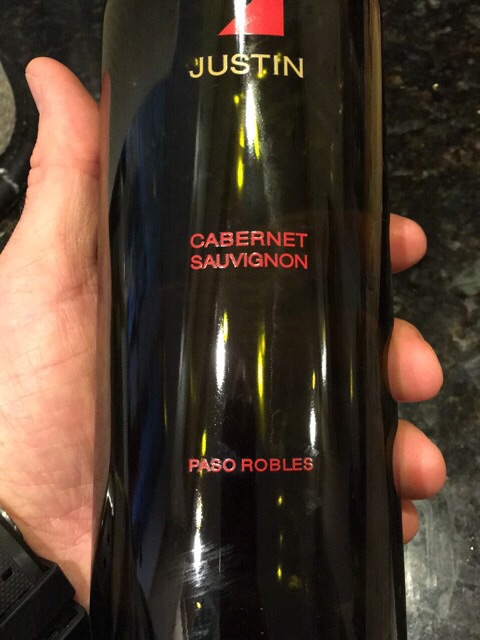}
      \end{subfigure}
      \caption{\textbf{Examples of images.} The viewpoint, lighting, and composition vary across images. %As illustrated in the first row, the label design of a vintage can also vary, although this rarely happens.
      }
      \label{fig:image_examples}
  \end{figure}
    % information about the data
    \textbf{Annotated flavors.} The flavor data consists of over 5k human-annotated pairwise similarities between 108 unique bottlings. Each annotated pair is annotated at least five times to reduce noise. %ensure representative observations.
    % \fred{More we can say about the flavor data? Can we add a visualization or give some more insights into what the flavor data looks like? }

    % overall stats of experiment 
    These annotations are collected through a series of wine-tasting events attended by a total of \num{256} non-expert wine drinkers. Most participants were between 21-25 years old, and more than half of them were from Denmark. Each participant volunteered their time, dedicating a maximum of two hours to complete the annotations. The experiment was conducted in accordance with the ''De Videnskabsetiske Komiteer'' (e. the Danish ethics committee for science) (see \cref{sec:ethics}).

  % how the experiment was conducted
  We randomly selected 5 wines for the participants to taste. The participants did not have access to any information regarding the individual wines. The wine was poured into non-transparent shot glasses and the labels of the wines were covered during the entire experiment. The participants were instructed to put colored stickers (representing each of the five wines) on a sheet of paper based on their taste similarity, closer meaning more similar. The participants could repeat the process up to three times, ensuring they did not consume more than 225 ml of wine. The average participant repeated the experiment two times.

  % how we digialized
  We automatically digitized the participants' annotations by taking a photo of each filled-out sheet. We used the Harris corner detector \citep{harris1988combined} to find the corners of the paper and a homographic projection to obtain an aligned top-down view of the paper. The images were mapped into HSV color space and a threshold filter applied to find the different colored stickers that the participant used to represent the wines. Having identified the location, we computed the Euclidean pixel-wise distance between all pairs of points, resulting in a distance matrix of wine similarities. A more detailed description of the collection and digitization of the napping papers can be found in \ref{sec:data_collection_details}.

\begin{figure}[!htb]
    \centering
    \begin{subfigure}[b]{0.32\textwidth}
        \includegraphics[width=\textwidth]{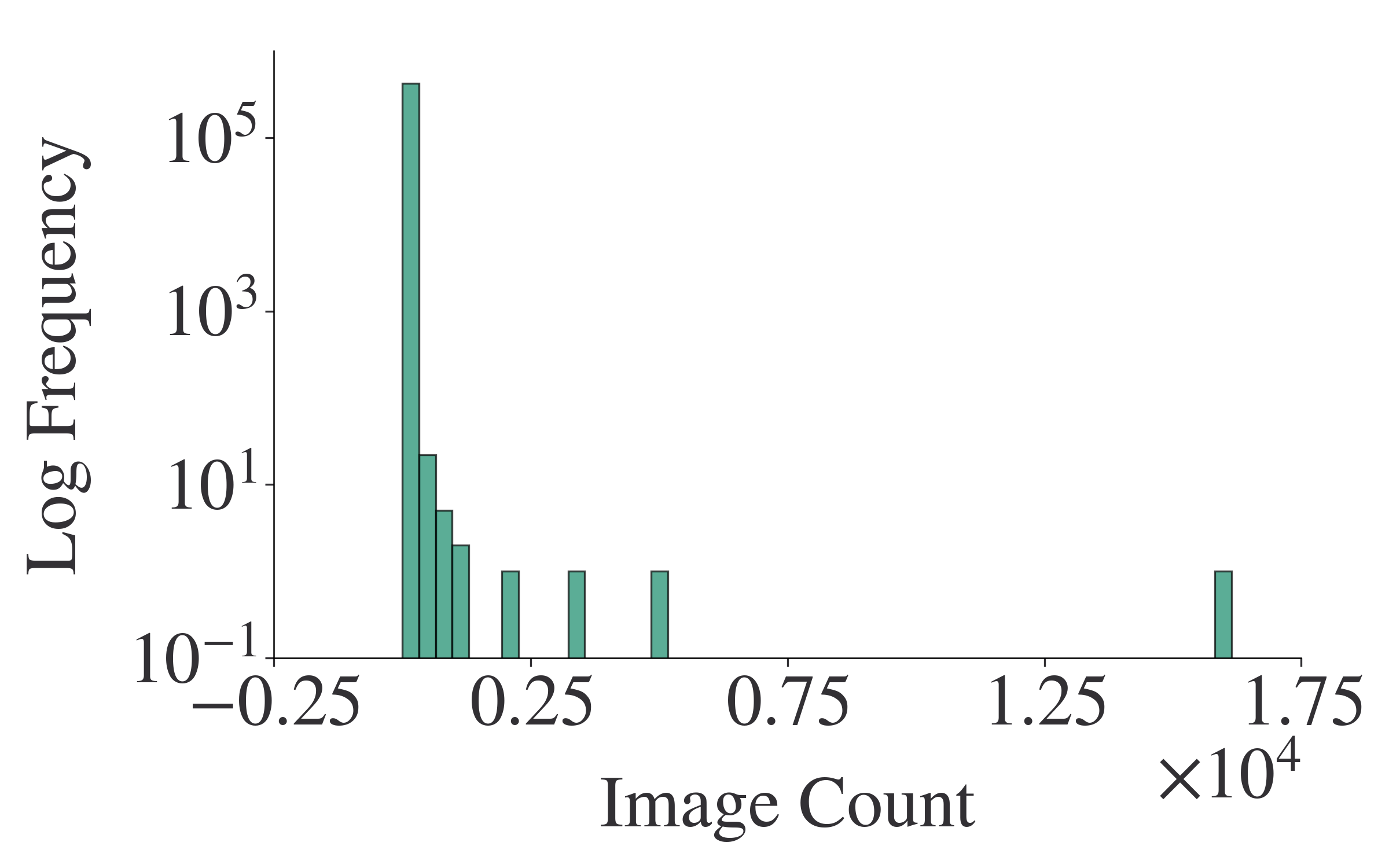}
        \caption{Images per unique bottling}
        \label{fig:images}
        \end{subfigure}
        \begin{subfigure}[b]{0.32\textwidth}
        \includegraphics[width=\textwidth]{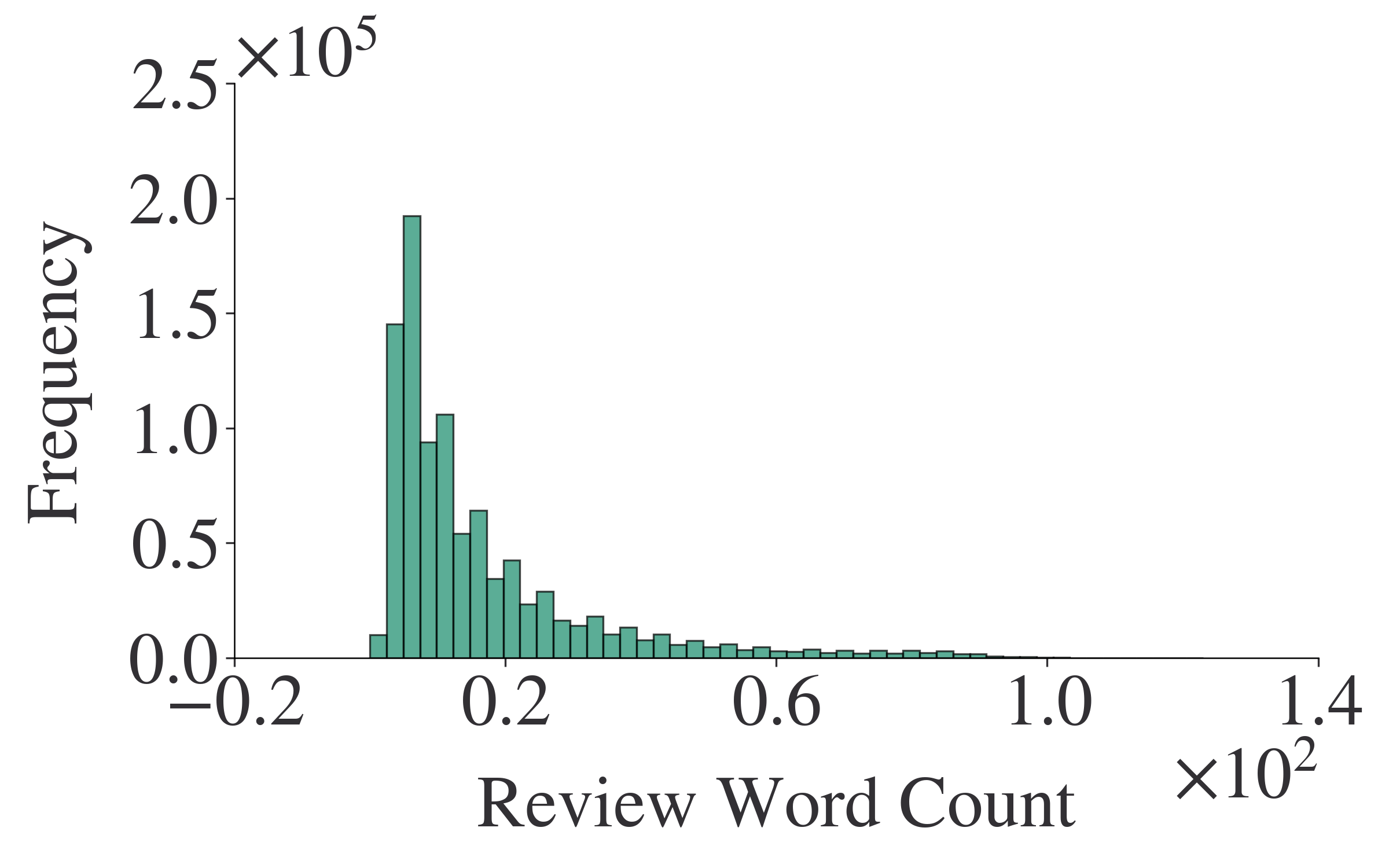}
        \caption{Words per review}
        \label{fig:reviews}
      \end{subfigure}
      \begin{subfigure}[b]{0.32\textwidth}
        \includegraphics[width=\textwidth]{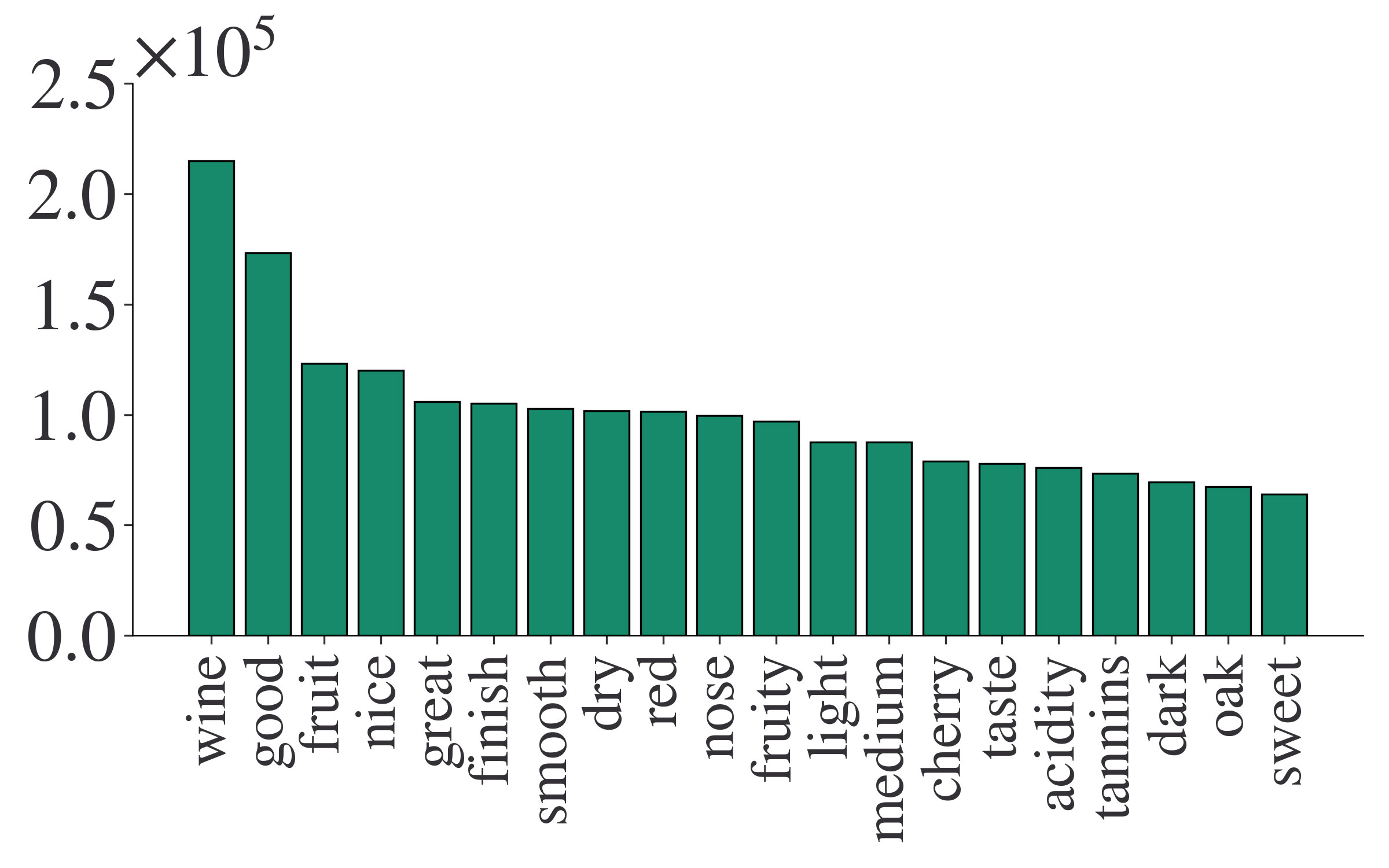}
        \caption{Most common keywords}
        \label{fig:words}
      \end{subfigure}
      \caption{\textbf{Summary statistics of user reviews and images.} Most unique bottlings have less than 10 images. The average review length is 16 words. Common keywords in the reviews include `fruit', `dry', and `smooth' revealing coarse semantic information about the flavor of the wines while other keywords such as `good' and `great' do not reveal flavor information. }
      \label{fig:user-image-reviews}
    \end{figure}

  \textbf{User-reviews.} We curated 824k text reviews from the Vivino platform. The reviews were filtered to contain at least \num{10} characters to avoid non-informative reviews such as `good' and `bad.' \cref{fig:datasetoverview} shows examples of user-reviews. The reviews are free text and can contain special tokens such as emojis. The reviews tend to describe price, pairing, and general terms of wine. Some also describe which flavors the reviewer tastes. These reviews are subjective and can vary based on personal factors and context, leading to inconsistent flavor profiles. Moreover, they only contain coarse flavor descriptions and focus more on aspects like preference, price, occasion, and so forth. \cref{fig:user-image-reviews} shows the distribution of word count per review, number of reviews per unique bottling, and the most common keywords.
  
\textbf{Images.}
  The dataset has 897k images of wine labels. Wine labels are known to play a major role in a consumer's decision to purchase a particular wine, so it is reasonable to believe that label design carries information regarding the taste of the wine \citep{talbot:forbes:2019}. \cref{fig:image_examples} shows examples of images from the dataset. The images vary in their viewing angle, illumination, and image composition. %In addition, vintages are not guaranteed to always have the same design, making it challenging to extract information about taste or identify similar vintages as seen the 6th image in the first row. \fred{Probably need to double check that this is correct. An observation from Simon's thesis, but I haven't looked into it myself. Seems a bit strange...}

\begin{figure}
  \centering
  \begin{subfigure}{0.33\textwidth}
    \centering
    \includegraphics[width=\linewidth]{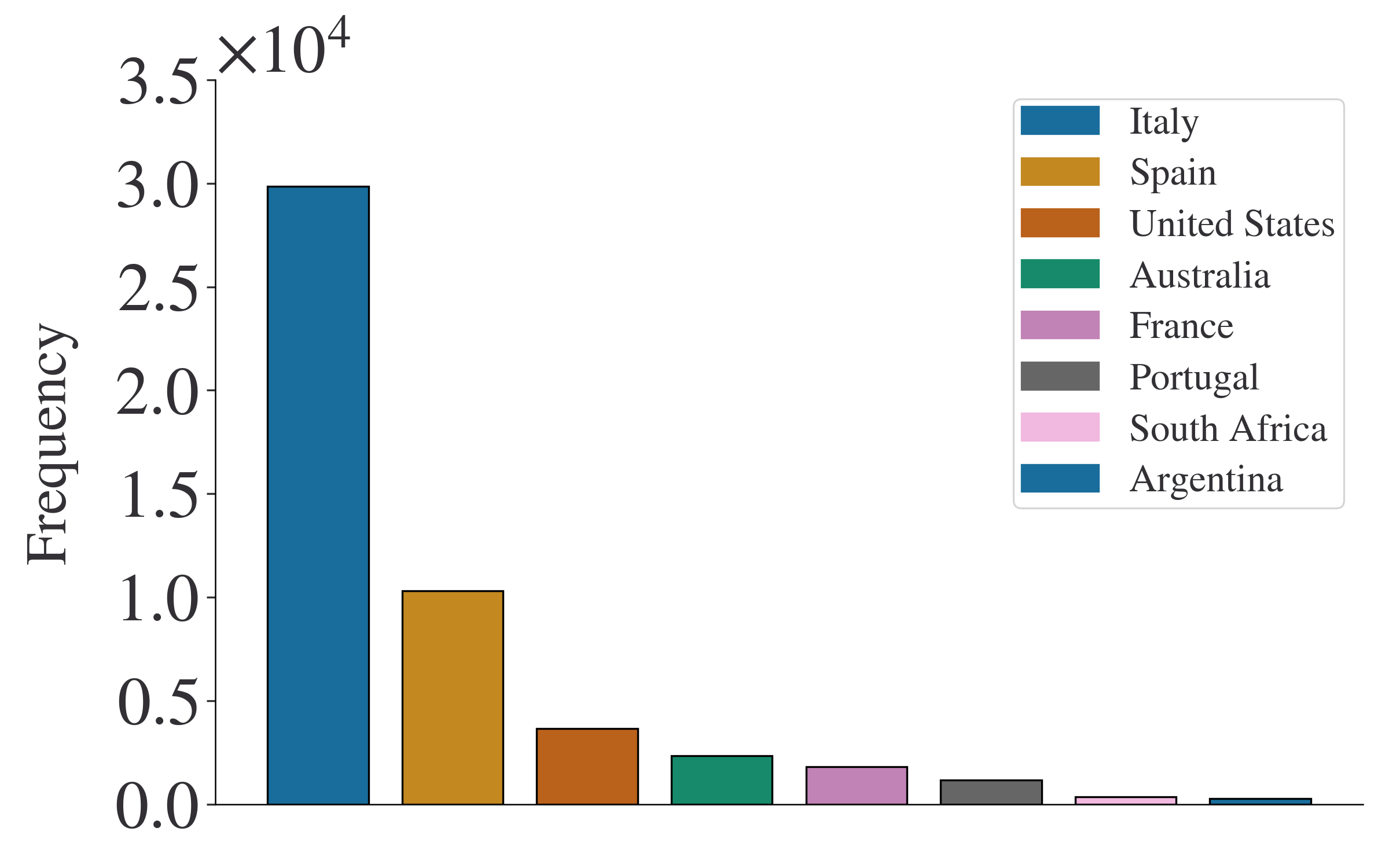}
    \caption{Country}
    \label{fig:country}
  \end{subfigure}%
  \begin{subfigure}{0.33\textwidth}
    \centering
    \includegraphics[width=\linewidth]{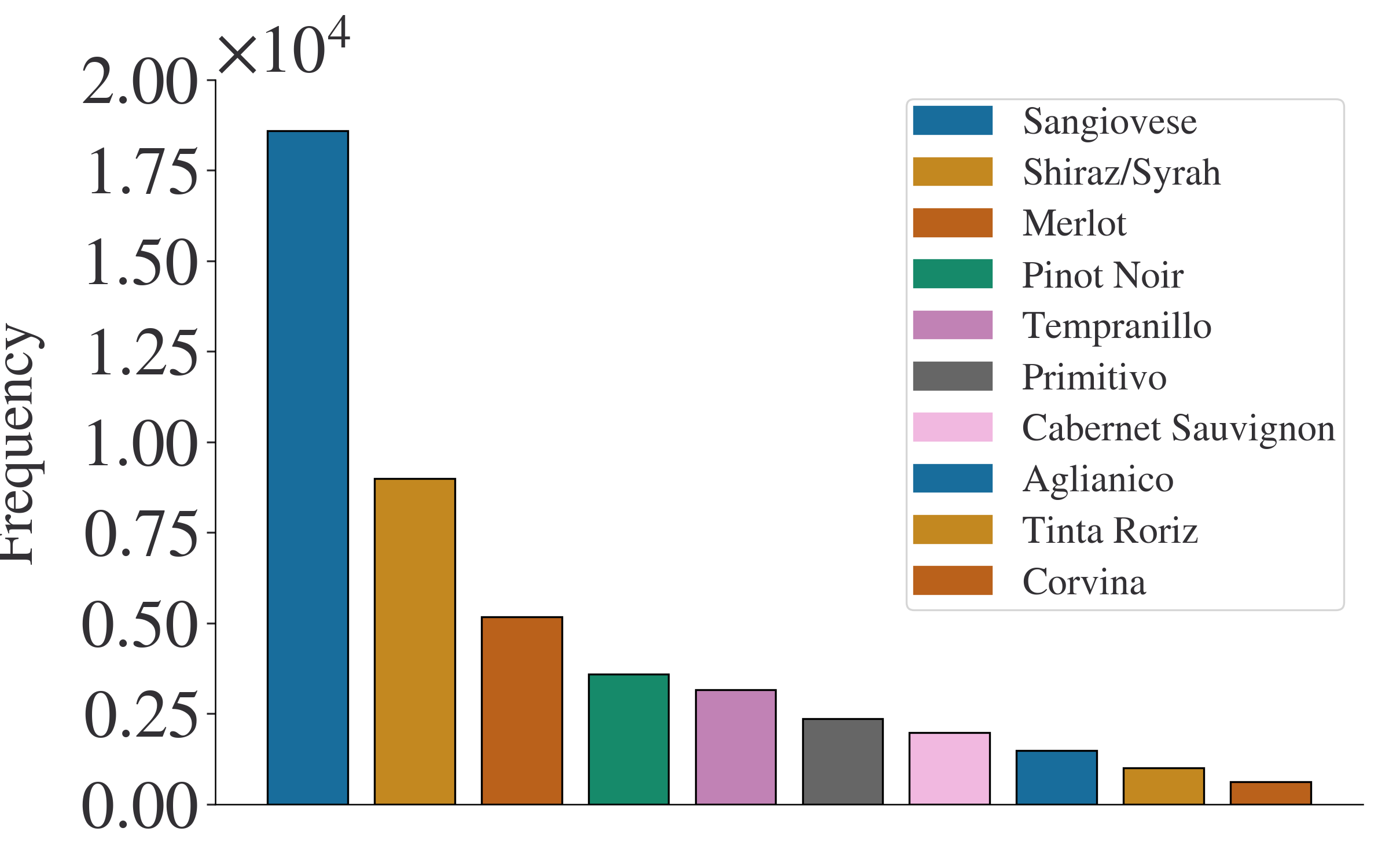}
    \caption{Grape}
    \label{fig:grape}
  \end{subfigure}%
  \begin{subfigure}{0.33\textwidth}
    \centering
    \includegraphics[width=\linewidth]{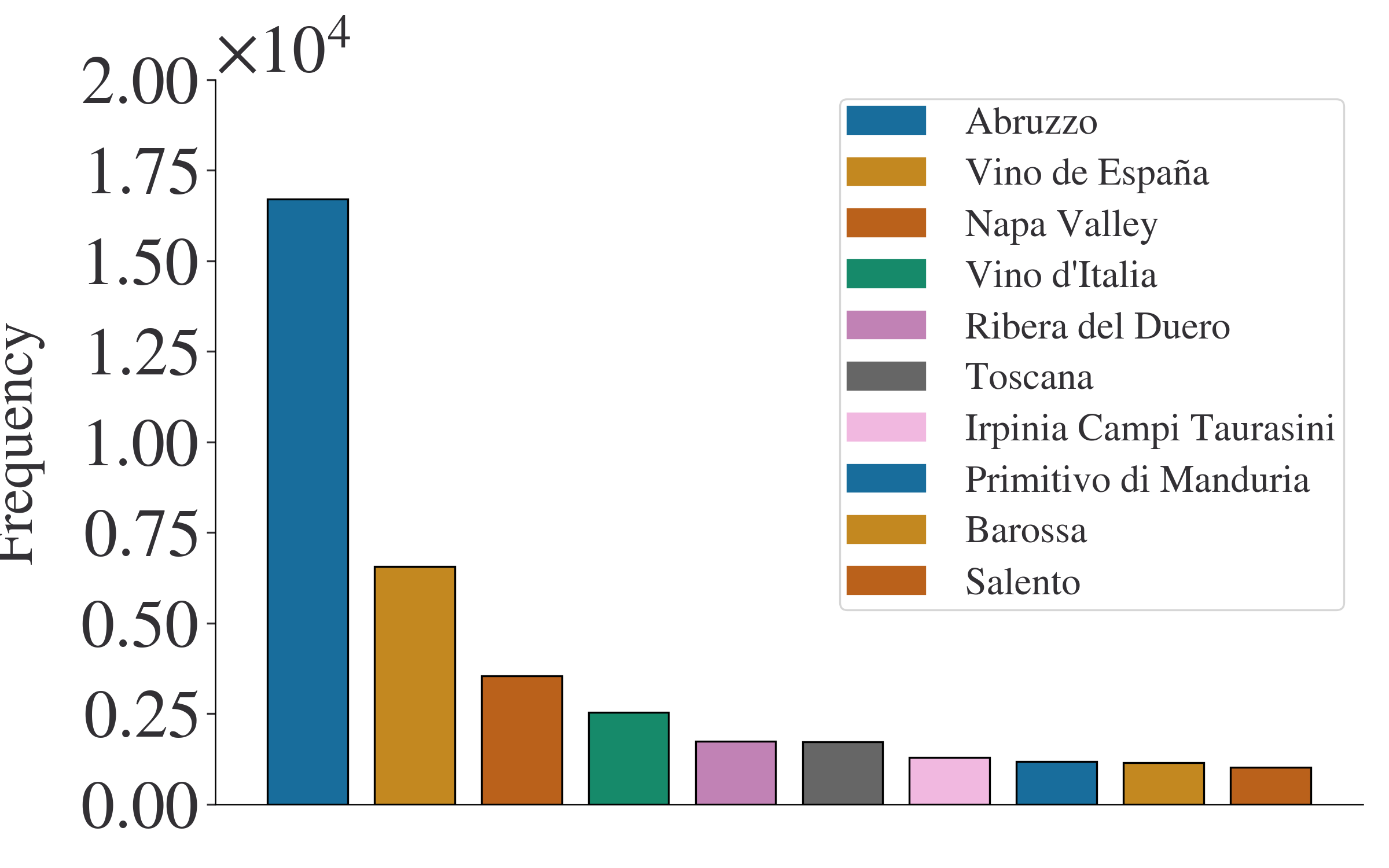}
    \caption{Region}
    \label{fig:region}
    \end{subfigure}%
    \\
  \begin{subfigure}{0.24\textwidth}
    \centering
    \includegraphics[width=\linewidth]{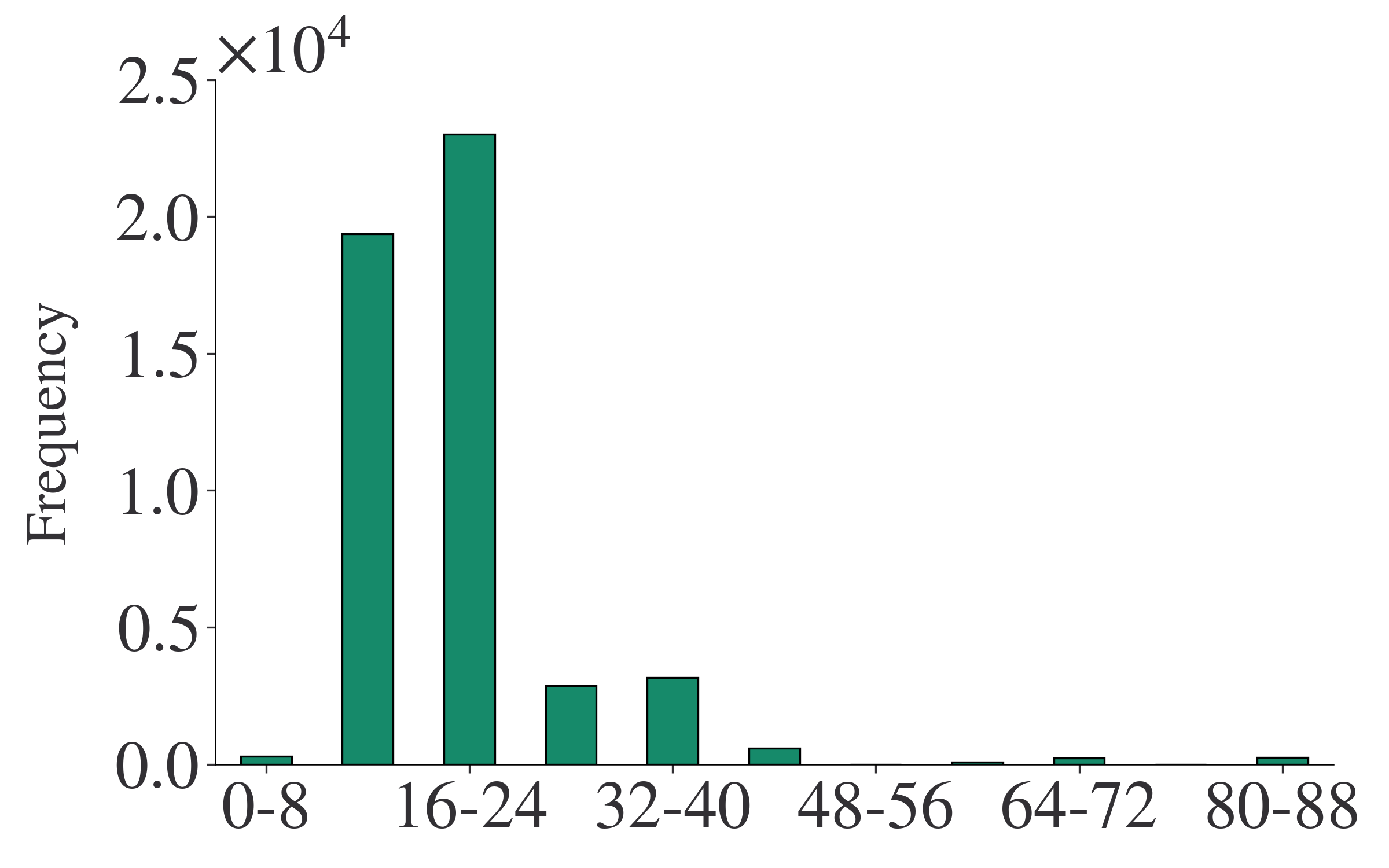}
    \caption{Price}
    \label{fig:price}
  \end{subfigure}
  \begin{subfigure}{0.24\textwidth}
    \centering
    \includegraphics[width=\linewidth]{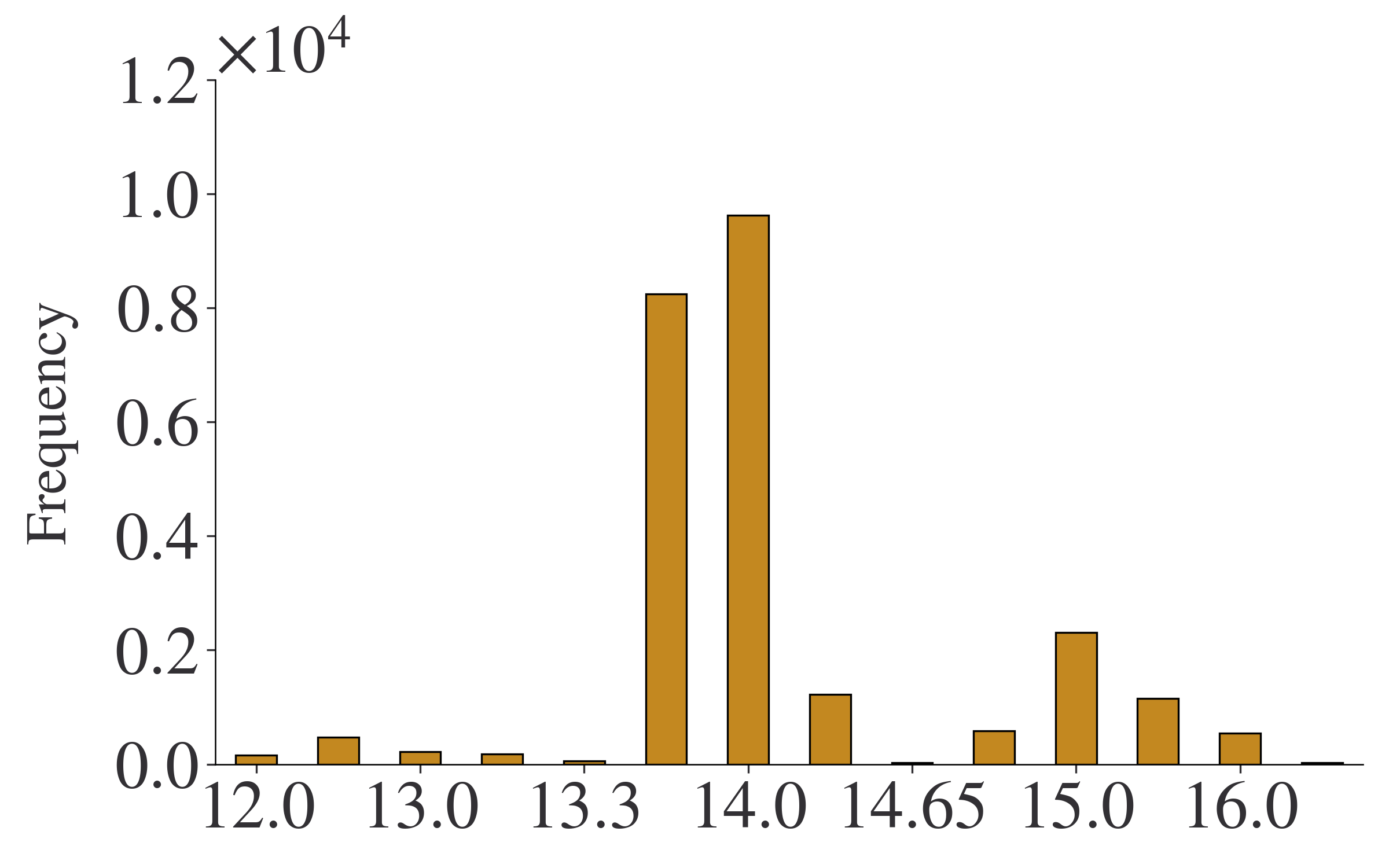}
    \caption{Alcohol content}
    \label{fig:alchol}
  \end{subfigure}%
  \begin{subfigure}{0.24\textwidth}
    \centering
    \includegraphics[width=\linewidth]{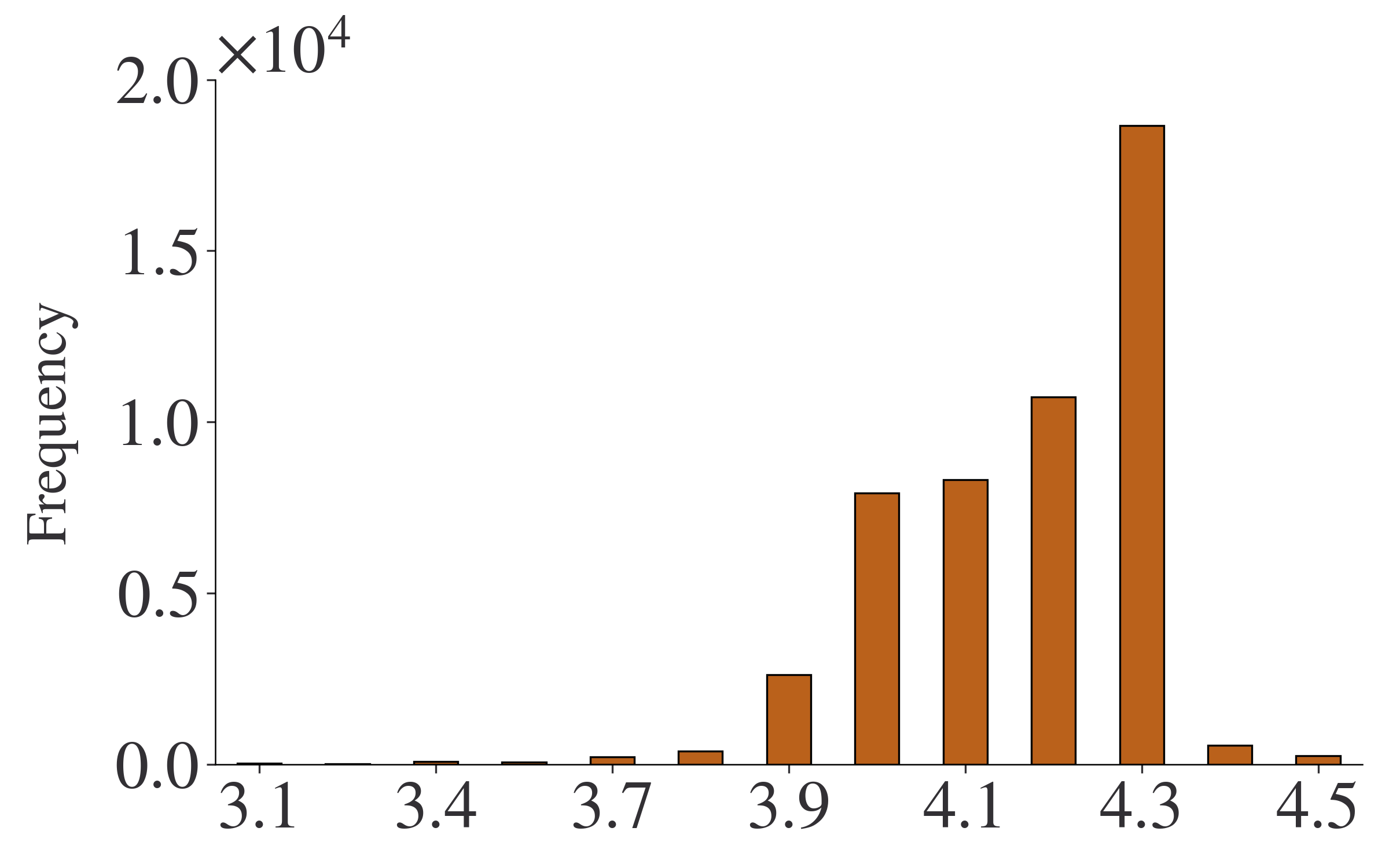}
    % \vspace{-0.2cm}
    \caption{Rating}
    \label{fig:rating}
  \end{subfigure}%
  \begin{subfigure}{0.24\textwidth}
    \centering
    \includegraphics[width=\linewidth]{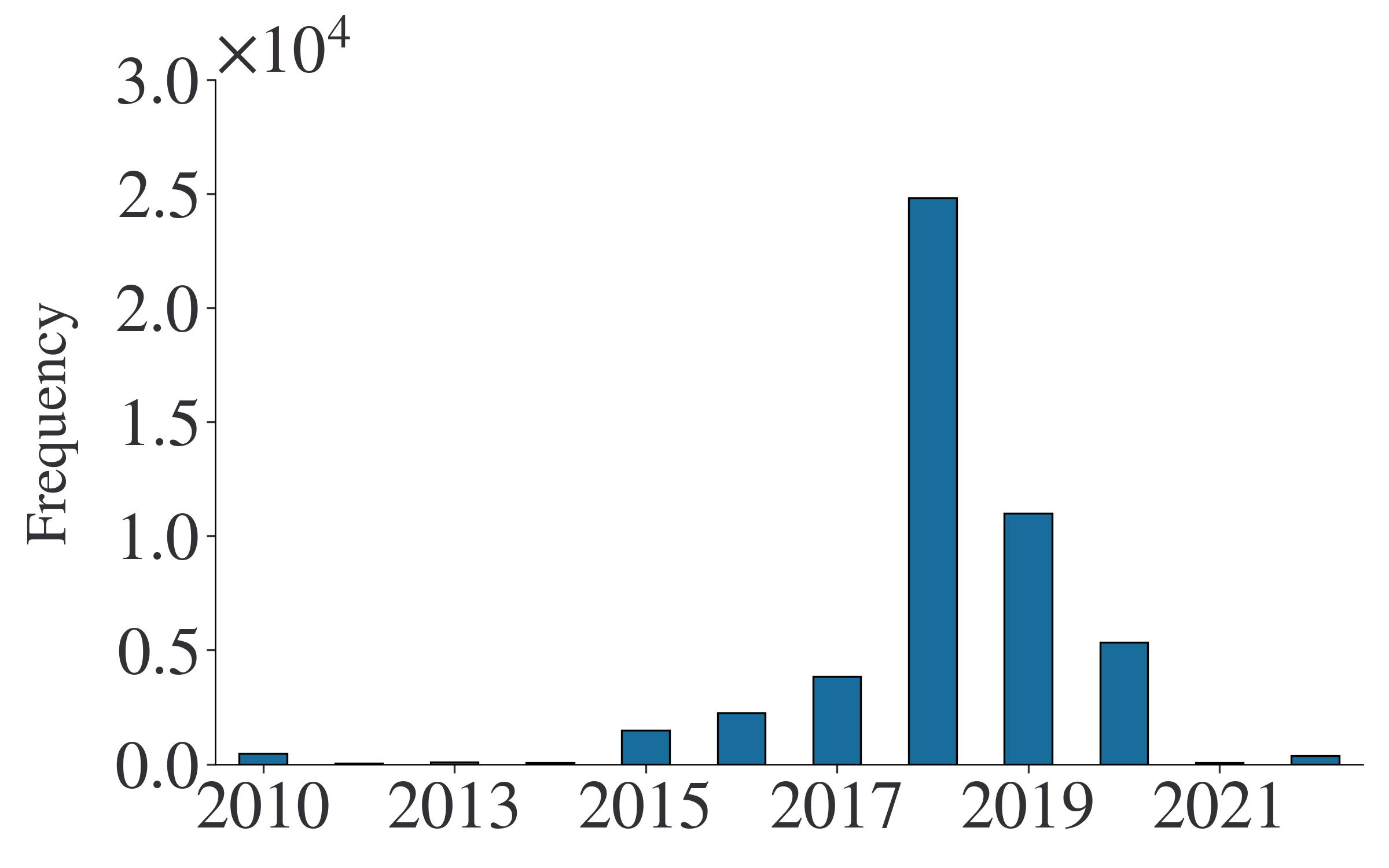}
    \caption{Year}
    \label{fig:year}
  \end{subfigure}
  \caption{\textbf{Wine attributes.} WineSensed contains attributes about the geolocation of production (country, region) and the grape composition of each wine. Furthermore, the dataset includes information on the average price of the wine, alcohol percentage, average rating on the Vivino platform, and the year of production. The histograms show the distribution of these attributes. }
  \label{fig:distribution_meta_data}
\end{figure}

  \textbf{Attributes.} Each wine is associated with the geographical location of the vineyard (both country and region), grape varietal composition, vintage, alcohol content, pricing, and average user rating. \cref{fig:distribution_meta_data} shows the distribution of these attributes. Most wines originate from Italy, with Sangiovese being the most commonly used grape. The wines occupy the lower range of the price spectrum, with the most expensive ones priced at around 40 USD. The attributes are available for 5\% of the dataset entries.

  % \begin{figure}
  %    \centering
  %    \includegraphics[width=\textwidth]{images/figure3 (1).png}
  %    \caption{\textbf{Meta-data.} Distribution of the meta data of the wines in the dataset. }
  %    \label{fig:distribution_meta_data}
  %\end{figure}

\section{Flavor Embeddings from Annotated Similarity \& Text-Image (FEAST)}

\begin{figure*}
    \centering
    \includegraphics[width=\textwidth]{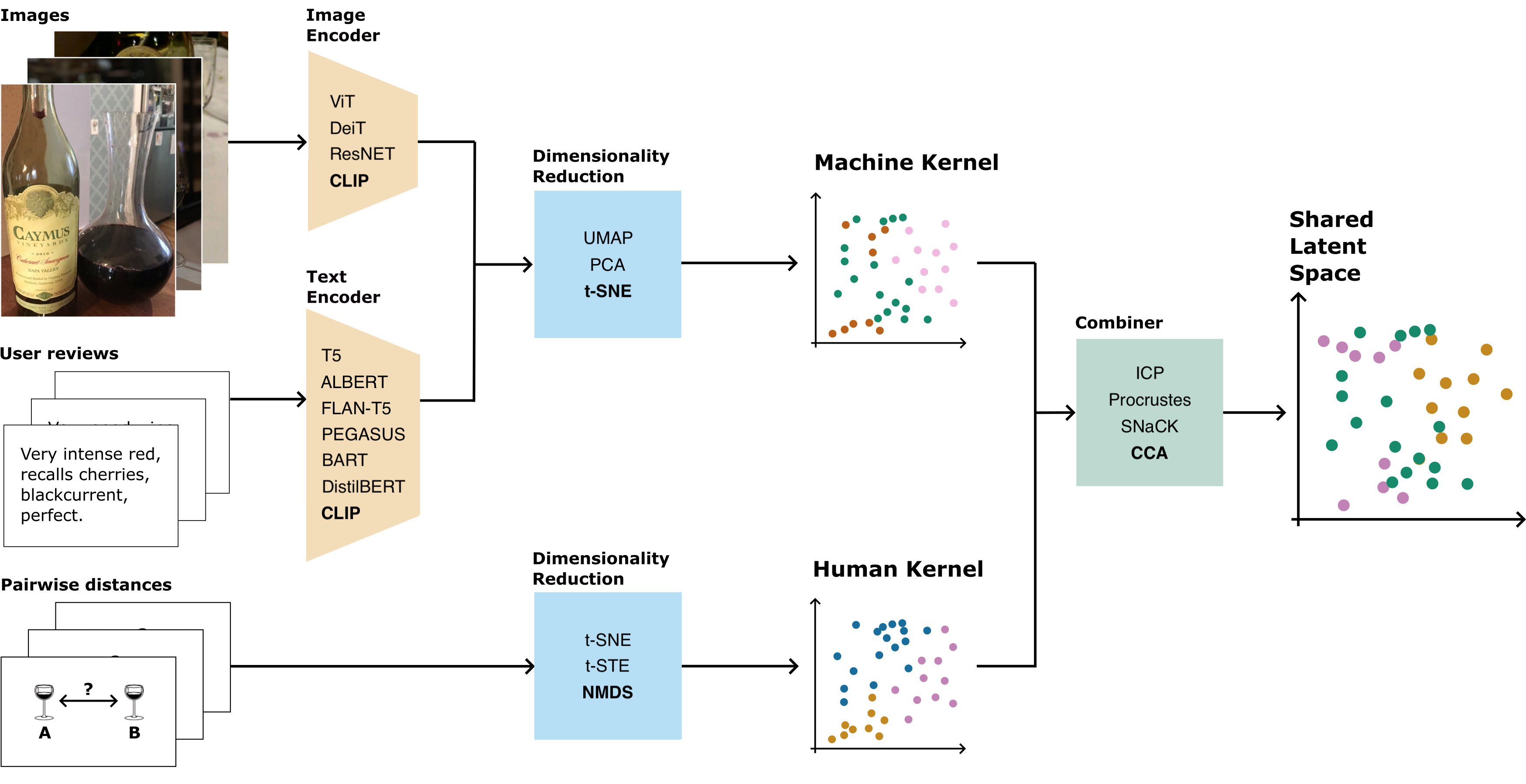}
    \caption{\textbf{Model overview.} FEAST takes text and/or images as input as well as human-annotated flavor similarities. The text and/or images are embedded into a latent representation with CLIP. We use NMDS to embed the flavor similarities. The two representations are aligned with CCA to produce a latent space that uses the structural information in CLIP embeddings and the intricacies of human annotations. The bolded methods in the orange, blue, and green boxes indicate choices for our best model, and their remaining combinations serve as an overview of the evaluated baselines. }
    \label{fig:model_overview}
\end{figure*}

The embeddings of recent large image and text networks contain structural and semantic information, however, they do not model the intricacies of human flavor. We propose FEAST, a method to align these embeddings to the human perception of flavor using a small set of human-annotated flavor similarities. FEAST takes text and/or images as input, as well as human-annotated flavor similarities. It outputs a unified embedding that aligns with human sensory perception. \cref{fig:model_overview} provides an overview of the proposed method.

%Our method has wide-ranging applications in recommendation systems, comparison, and classification of multimodal datasets. I think a sentence like this belong in the introduction. 

%\paragraph{Embedding Modalities with CLIP.} 
We first embed the text and/or images into a latent space with CLIP \citep{DBLP:journals/corr/abs-2103-00020}. We use CLIP because of its large training corpus and its image-text aligned latent space, however, highlights that other pretrained networks can be used. We use t-SNE \citep{van2008visualizing} to reduce the dimensionality of the latent space to 2, which simplifies and constrain the later alignment with the pairwise flavor annotations.

%The pairwise flavor annotations are represented as a sparse matrix, where empty elements represent missing relative distances. 
The pairwise distances are embedded into a 2D representation using Non-metric multidimensional scaling (NMDS) with the SMACOF strategy \citep{JSSv031i03}. NMDS allows us to preserve the original flavor distances provided by humans in a shared space, where each unique bottling is represented with point location, rather than pairwise distances. MDS is commonly used in food science to analyze sensory annotations from Napping studies \citep{pineau2022comparison, varela2012sensory, nestrud2010perceptual}.

% \fred{I think we should also highlight the relation from MDS to PCA.}

% \thoranna{This could be a good place to cite food science papers that use MDS to process napping paper results}\fred{YES!}

We then align these two 2D representations to get a joint representation that benefits from the structural and semantic information of the image and/or text representations, scales to unobserved unique bottlings, and is aligned with the human perception of flavor. We use Canonical Correlation Analysis (CCA) \citep{harold1936relations} to align the two representations. CCA identifies and connects common patterns between these representation spaces, ensuring that the final representation is consistent across all input modalities.

\section{Experiments}
We conduct two experiments on the WineSensed dataset. First, we explore how well recent large pretrained language and image models explain wine attributes that correlate with the flavor of a wine. Second, we explore multimodal models' capabilities to represent more intricate flavors. %Also in this setting, we find that combining large pretrained models with human flavor annotations improves flavor predictions over using a single modality model. %We selected the reviews, images, and attributes associated to the 108 wines that were annotated during our data collection events for the WineSensed dataset.
% As \cref{fig:user-image-reviews}\subref{fig:images} shows, the number of images associated with each wine varies quite significantly, with the majority having fewer than 10 images, and only a small portion associated with thousands of images. To standardize our process and also optimize computational efficiency, we decided to utilize a random sampling strategy, limiting the selection to a maximum of 100 images for any given wine.

%We find that in most cases, both language and image models' predictions improve when aligned with human flavor annotations. In fact, we find that optimal classification results are achieved by combining images, text, and flavor annotations, highlighting the usefulness of the proposed multimodal dataset.

\textbf{Experimental setup.} We explore several configurations of human kernels, machine kernels, and ``combiners'' that align the two representations. \cref{fig:model_overview} provides an overview of our baselines. The \textbf{human kernel} is formed with t-STE \citep{van2012stochastic}, a low dimensional graph representation reduced with t-SNE or NMDS, where the notable difference is that t-STE discards the flavor distances, and solely optimizes for triplet orderings. The \textbf{machine kernel} consists of two steps: (1) we use a pretrained model to embed text and/or images into a low dimensional space, (2) which is then compressed into a two-dimensional space. For (1), we explore DistilBert \citep{sanh2019distilbert}, T5 \citep{raffel2020exploring}, ALBERT \citep{lan2019albert}, BART \citep{lewis2019bart}, PEGASUS \citep{zhang2020pegasus}, FLAN-T5 \citep{chung2022scaling} and CLIP for embedding text and ViT \citep{dosovitskiy2020image}, ResNet \citep{he2016deep}, DeiT \citep{touvron2021training}, and CLIP for embedding images. For (2), we explore t-SNE, UMAP \citep{mcinnes2018umap}, and PCA \citep{pearson1901liii}. For the \textbf{combiners}, we experiment with CCA, Iterative Closest Point (ICP) \citep{chen1992object}, Procrustes \citep{gower1975generalized} and SNaCK. For a more detailed description of the implementation and software packages used, please refer to \ref{sec:training_details} the Appendix.

\subsection{Coarse flavor predictions}

We first explore how well pretrained language and vision models explain wine attributes that correlate with flavor. We then investigate if using FEAST to align the machine and human kernels improves the representation.

\textbf{Implementation details.} We use a balanced SVM classifier with an RBF kernel as well as a Multi-layer Perceptron \citep{} neural network to predict wine attributes of the flavor embeddings. We predict price, alcohol percentage, rating, region, country, and grape variety as these attributes are known to correlate with the perceived wine flavor. We mitigate imbalanced class distributions with class weight balancing and oversampling of the minority classes. We report the accuracy averaged over the seven attributes computed through 5-fold cross-validation. The accuracy measures how coherent the embeddings are with the flavor attributes. A more detailed description of the implementation can be found in \ref{sec:more_classification_results}.

\begin{table}[t!]
    \small
    \begin{minipage}[t]{0.37\textwidth}
    \centering
    \caption{\textbf{Ablation of machine kernels.} Accuracy of machine kernels across image and text modalities. Image models perform worse than text models. ALBERT, BART and CLIP perform the best, all models perform better than random using at least one classification method.
    }
    \scalebox{0.85}{
    \begin{tabular}{ll|ll} \toprule
        \multicolumn{2}{c|}{} & \multicolumn{2}{c}{Acc $\uparrow$} \\
Machine kernel & Modality & SVM & NN \\ \midrule
        Random & & 0.11 & 0.11 \\
        ViT & Image & 0.09 & 0.13 \\
        DeiT & Image & 0.14 & 0.15 \\
        ResNET & Image & 0.15 & 0.16 \\
        CLIP & Image & 0.11 & 0.15 \\
        T5 & Text & 0.15 & 0.16 \\
        \textbf{ALBERT} & \textbf{Text} & 0.15 & \textbf{0.18} \\
        \textbf{BART} & \textbf{Text} & \textbf{0.16} & 0.15 \\
        DistilBERT & Text & 0.15 & 0.17\\
        \textbf{CLIP} & \textbf{Text} & \textbf{0.16} & \textbf{0.18} \\
        FLAN-T5 & Text & 0.15 & 0.17 \\
        PEGASUS & Text & 0.13 & 0.13 \\
        BART & Text & 0.11 & 0.15 \\
        \bottomrule
    \end{tabular}
    }
    \label{tab:pre_trained}
    \end{minipage}
    \hspace{0.1cm}
    \begin{minipage}[t]{0.29 \textwidth}
    \caption{\textbf{Ablation of Modalities.} Accuracy of single and combined modalities. Using multiple modalities improves performance. We find that combining image, text, and flavor yields much better accuracy than modeling each modality separately.} \vspace{1.38cm}
    \scalebox{0.85}{
    \begin{tabular}{l|ll} \toprule
    \multicolumn{1}{c|}{} & \multicolumn{2}{c}{Acc $\uparrow$} \\
    Modality & SVM & NN \\ \midrule
    Flavor & 0.16 & 0.11 \\
    Image & 0.11 & 0.15 \\
    Text & 0.16 & 0.18 \\
    Text+Flavor & 0.23 & 0.18 \\
    Image+Text & 0.22 & 0.25 \\
    Image+Flavor & 0.23 & 0.18 \\
    \textbf{Image+Text+Flavor} & \textbf{0.28} & \textbf{0.26} \\
    \bottomrule
    \end{tabular}
    }
    \label{tab:modality}
    \end{minipage}
    \hspace{0.35cm}
    \begin{minipage}[t]{0.25 \textwidth}
    \centering
    \caption{\textbf{Ablation of human kernels, reducers, and combiners.}} \vspace{0.6cm}
    \scalebox{0.85}{
    \begin{tabular}{p{2cm}|ll} \toprule
        \multicolumn{1}{c|}{Reducer} & \multicolumn{2}{c}{Acc $\uparrow$} \\
    Human Kernel & SVM & NN \\ \midrule
        Random & 0.11 & 0.11 \\
        t-STE & 0.13 & 0.10 \\
        t-SNE & 0.15 & 0.13 \\
        \textbf{NMDS} & \textbf{0.16} & \textbf{0.13} \\
        %\bottomrule
    \end{tabular}
    }
    %\caption{ \thoranna{Which reducer is best. fixed:model=CLIP(images+text)}}
    \scalebox{0.85}{
    \begin{tabular}{p{2cm}|ll} \midrule
    \multicolumn{1}{c|}{Reducer} & \multicolumn{2}{c}{Acc $\uparrow$} \\
    Machine Kernel & SVM & NN \\ \midrule
    UMAP & 0.15 & 0.18 \\
    PCA & 0.20 & 0.21 \\
    \textbf{t-SNE} & \textbf{0.22} & \textbf{0.25} \\
    %\bottomrule
    \end{tabular}
    }
    %\caption{ \thoranna{Which reducer is best. fixed:model=CLIP(images+text)}}
    \scalebox{0.85}{
    \begin{tabular}{p{2cm}|ll} \midrule
    \multicolumn{1}{c|}{} & \multicolumn{2}{c}{Acc $\uparrow$} \\
    Combiner & SVM & NN \\ \midrule
    ICP & 0.21 & 0.24 \\
    Procrustes & 0.19 & 0.23 \\
    SNaCK & 0.23 & 0.24 \\
    \textbf{CCA} & \textbf{0.28} & \textbf{0.26} \\
    \bottomrule
    \end{tabular}
    }
    \label{tab:performance_comp}
    \end{minipage}
\end{table}

% Figure \ref{fig:classification_results}\subref{fig:averages} shows that the average classification accuracy using combined embeddings outperforms random guessing in 6 out of 7 classes. In addition, merging these embeddings improves accuracy over Machine Kernel and Human Kernel, respectively, in 5 out of 7 classes. The average accuracy using combined embeddings is 18.09\%, which is significantly higher than the 9.15\% achieved with random guessing.
% With the use of combined embeddings, we see an average accuracy improvement of 92.19\%. However, relying on random guessing results in an average accuracy \textit{decrease} of 163.74\%

 % Table \ref{tab:class_res} shows that both the text and the image embeddings yields higher classification accuracy than the random baseline for 6 out of 7 classes, suggesting that both these machine kernels bears useful information to describe wine flavor. Furthermore, we find that combining the machine kernel with the human kernel improves the accuracy over both the machine and human kernel in 5 out of 7 classes. The average accuracy using the combined embeddings is 18.09\%, notably higher than the 9.15\% achieved with random guessing\fred{here I compare to the human and machine kernel rather than random guessing, e.g. "notably hugher than teh XX percent found with the machine kernel and XX percent found with the human kernel}. This shows that combining the embeddings leads to useful representations that correlate to concepts known to describe the taste of the wine.

\textbf{Results.} \cref{tab:pre_trained,tab:performance_comp,tab:modality} ablates our proposed method and summarizes our main conclusions. Please see \cref{sec:more_classification_results} for per attribute classification accuracy for all combinations of machine kernels, human kernels, modalities, reduces, and combiners. 

\cref{tab:pre_trained} shows that most pretrained image and text models yield slightly higher  performance than the random baseline. The text encoders are slightly better than the image encoders. BART and CLIP perform the best. All encoders in the table use t-SNE to reduce the embedding to 2D. \cref{tab:performance_comp} (middle) shows t-SNE yields better accuracy than UMAP and PCA when using a CLIP encoder.

\cref{tab:performance_comp} (top) shows that NMDS performs better than t-STE. NMDS uses the relative distances between annotations, whereas t-STE discretizes the annotations and considers only the ordering within each triplet. The results suggest that the pairwise distances are useful to model the flavor space. \cref{tab:performance_comp} (bottom) shows that using CCA to align the two representations yields higher accuracy than SNaCK or ICP. 

\cref{tab:modality} shows that including flavor as a modality increases the accuracy, \textit{e.g.} using flavor to align the image or text embeddings lead to higher accuracy. Using CLIP followed by t-SNE, NMDS, and CCA to combine language, vision, and flavor into a single representation leads to the best configurations, illustrating that the human annotations are useful for learning a flavor representation. Maybe most surprisingly, we show that each modality by itself is on par with the random baseline, but their combination produces a latent space that much better describes the flavor attributes.

 \subsection{Fine-grained flavor predictions}
We now proceed to evaluate more intricate flavor predictions by using human-annotated flavor similarities as ground truth.

\textbf{Implementation details.} To evaluate our representation, we measure the Triplet Agreement Ratio (TAR) \citep{6349720} between our predicted flavor embeddings and the human-annotated flavors. TAR measures the agreement between a triplet derived from the latent space and the ground truth triplets from the flavor annotations. Higher TAR means that the ordering of distances in the latent space corresponds to the human perception of flavor. This measure indicates how aligned the two representations are, and provides a higher granularity of flavor prediction than flavor attributes. A more detailed description of the implementation can be found in \ref{sec:more_results}.

\textbf{Results.} \cref{tab:summary} ablates FEAST and shows that for the higher granularity predictions both the pretrained text and image encoders improve upon the random baseline. We show that including the human kernel with NMDS further improves the TAR scores. This highlights the usefulness of the flavor distances recorded by the human annotators. In \cref{sec:more_results}, we show results from all configurations of human kernels, machine kernels, reducers, and combiners. We find that NMDS consistently yields better performance than t-STE, and that combining human and machine kernels improves the TAR scores across multiple model configurations.

\begin{table}[]
    \centering
    \small
    \caption{\textbf{Fine-grained flavor predictions.} Triplet Agreement Ratio (TAR) between text, image, and multi-modal encoders and human annotated flavor similarities. A higher TAR indicates that the model's representation space is more aligned with humans' perception of flavor.}
    \begin{tabular}{lll|ll} \toprule
        Machine Kernel & Human Kernel & Combiner & Modality & TAR $\uparrow$ \\ \midrule
        % T5 + UMAP & & & Text only & 0.82 \\
        % T5 + t-SNE & & & Text only & 0.82 \\
        % T5 + UMAP & MDS & CCA & Text + flavor & 0.89 \\
        % T5 + t-SNE & MDS & ICP & Text + flavor & \textbf{0.90} \\
        % T5 + t-SNE & MDS & CCA & Text + flavor & \textbf{0.90} \\
        % T5 + UMAP & t-STE & CCA & Text + flavor & 0.83 \\
        % T5 + t-SNE & t-STE & ICP & Text + flavor & 0.78 \\
        % T5 + t-SNE & t-STE & SNaCK & Text + flavor & 0.84 \\
        % \midrule
        % DeiT + UMAP & & & Image only & 0.82 \\
        % DeiT + t-SNE & & & Image only & 0.83 \\
        % DeiT + UMAP & MDS & CCA & Image + flavor & 0.91 \\
        % DeiT + t-SNE & MDS & ICP & Image + flavor & 0.90 \\
        % DeiT + t-SNE & MDS & CCA & Image + flavor & \textbf{0.92} \\
        % DeiT + UMAP & t-STE & CCA & Image + flavor & 0.82 \\
        % DeiT + t-SNE & t-STE & ICP & Image + flavor & 0.78 \\
        % DeiT + t-SNE & t-STE & SNaCK & Image + flavor & 0.86 \\
        % \midrule
        Random & & & & 0.5 \\
        CLIP + t-SNE & & & Text & 0.82 \\
        CLIP + t-SNE & & & Image & 0.82 \\
        % CLIP + UMAP & & & Image + text & 0.82 \\
        CLIP + t-SNE & & & Image + Text & 0.81 \\
        CLIP + t-SNE & & & Image + Flavor & 0.89 \\
        CLIP + t-SNE & & & Text + Flavor & 0.88 \\
        % CLIP + UMAP & MDS & CCA & Image + text + flavor & \textbf{0.91} \\
        % CLIP + t-SNE & MDS & ICP & Image + text + flavor & 0.90 \\
        CLIP + t-SNE & NMDS & CCA & Image + Text + Flavor & \textbf{0.91} \\
        % CLIP + UMAP & t-STE & CCA & Image + text + flavor & 0.84 \\
        % CLIP + t-SNE & t-STE & ICP & Image + text + flavor & 0.78 \\
        % CLIP + t-SNE & t-STE & SNaCK & Image + text + flavor & 0.79 \\ \bottomrule
        \bottomrule
    \end{tabular}
    \label{tab:summary}
\end{table}

\section{Discussion \& Conclusion} \label{conclusion}

In this paper, we introduce WineSensed, an extensive multimodal dataset curated for flavor modeling. The dataset comprises over 897k images and 824k reviews, and has over 5k human-annotated pairwise flavor similarities, obtained via a sensory study involving 256 participants. We propose a simple algorithm, FEAST, to align semantic information from machine kernels with flavor similarities from human annotators in a shared flavor representation. We find that combining these modalities improves both coarse and fine-grained flavor predictions.

WineSensed further strengthens the collaboration between the food science and machine learning communities, introduces flavor as a modality in multimodal models, and serves as an entry point for the development of machine learning models for flavor analysis and potentially deepening our comprehension of wine flavors. The dataset and the proposed procedures open many interesting possibilities, such as using flavor to ground foundation models or extending the dataset with other modalities, such as chemical composition, or other food categories.

\textbf{Constraints and considerations.} The dataset serves as a novel first step to including human-annotated flavor in the array of modalities in multimodal models. Its current scope is constrained to a selected group of red wines, predominantly Italian ones. While this enables a more nuanced understanding of flavors within Italian wines, it may not represent the broader spectrum of red wines globally. Furthermore, the dataset's emphasis on wines prevalent in Western cultures highlights a geo-cultural bias. Expanding the dataset to encompass more diverse drink types from different cultures could provide a more comprehensive understanding of global flavor perception. Lastly, the Napping methodology is not immune to the influences of participants' backgrounds and experiences. Individual perceptions, shaped by personal histories, can introduce nuances in the data. Though leveraging non-expert wine drinkers for flavor annotations introduces subjectivity, this approach, inspired by common sensory study practices, broadens taste perspectives, enhances study accessibility, and offers commercial value, with multiple annotations per entry mitigating individual biases. Exploring a broader range of foods and beverages remains a valuable direction for future work.

%We use Napping to obtain flavor annotations, although providing detailed information about flavor similarities, might not scale to all foods.

% TODO: put in the final version
\textbf{Acknowlegements.} This work was supported by the Pioneer Centre for AI, DNRF grant number P1, and by research grant (42062) from VILLUM FONDEN. This project received funding from the European Research Council (ERC) under the European Union’s Horizon 2020 research and innovation programme (grant agreement 757360), as well as the Danish Data Science Academy (DDSA).

\newpage

 \bibliographystyle{plainnat}
\bibliography{references}

\newpage
\appendix
\section{Project webpage}\label{sec:project_page}
We provide a project webpage for the dataset that can be found here: \url{https://thoranna.github.io/learning_to_taste/}, which contains a link to the dataset and the code to reproduce our experiments. Additionally, we provide more examples from our dataset and images from the data collection. 

\section{License}
The WineSensed dataset is licensed under the Creative Commons Attribution-NonCommercial-NoDerivatives 4.0 International License. Details can be found here: \url{https://creativecommons.org/licenses/by-nc-nd/4.0/}.

\section{The WineSensed file structure}\label{sec:dataset_structure}
Our dataset is currently available  here: \url{https://data.dtu.dk/articles/dataset/WineSensed\_Learning\_to\_Taste\_A\_Multimodal\_Wine\_Dataset/23376560}. The dataset will be maintained on this site, which is hosted on a server run by the Technical University of Denmark.

WineSensed contains a \texttt{metadata.zip} file consisting of the files \texttt{participants.csv}, which contains information connecting participants to annotations in the experiment, \texttt{images\_reviews\_attributes.csv}, which contains reviews, links to images, and wine attributes, and \texttt{napping.csv}, which contains the coordinates of each wine on the napping paper, alongside information connecting each coordinate pair to the wines being annotated and the participant that annotated them. The \texttt{chunk\_<chunk num>.zip} folders contain the images of the wines in the dataset in \texttt{.jpg} format. 

\texttt{napping.csv} contains the following fields: 

\begin{itemize}
  \item \texttt{session\_round\_name}: session number during the \texttt{event\_name}, at most three sessions per event (maps to \texttt{experiment\_round} in \texttt{participants.csv})
  \item \texttt{event\_name}: name of the data collection event (maps to the same attribute in \texttt{participants.csv})
  \item \texttt{experiment\_no}: the serial number of the napping paper in the \texttt{session\_round\_name} in which it was collected (maps to \texttt{experiment\_no} in \texttt{participants.csv})
  \item \texttt{experiment\_id}: id of the wine annotated
  \item \texttt{coor1}: x-axis coordinate on the napping paper
  \item \texttt{coor2}: y-axis coordinate on the napping paper
  \item \texttt{color}: color of the sticker used
\end{itemize}

\texttt{participants.csv} contains the following fields: 

\begin{itemize}
\item \texttt{session\_round\_name}: session number during the \texttt{event\_name}, at most three sessions per event (maps to \texttt{experiment\_round} in \texttt{napping.csv})
  \item \texttt{event\_name}: name of data-collection event (maps to \texttt{event\_name} in \texttt{napping.csv})
  \item \texttt{experiment\_no}: the serial number of the napping paper in the \texttt{session\_round\_name} in which it was collected (maps to \texttt{experiment\_no} in \texttt{napping.csv})
  \item \texttt{round\_id}: round number (from 1-3)
  \item \texttt{participant\_id}: id the participant was given in the experiment
\end{itemize}

\texttt{images\_reviews\_attributes.csv} contains the following fields: 

\begin{itemize}
  \item \texttt{vintage\_id}: vintage id of the wine
  \item \texttt{image}: image link (each \texttt{<image name>.jpg} in \texttt{chunk\_<chunk num>.zip} can be mapped to a corresponding image link in this column by removing the \texttt{/p} prefix from the link).  
  \item \texttt{review}: user review of the wine
  \item \texttt{experiment\_id}: id the wine got during data collection (each \texttt{experiment\_id} can be mapped to the same column in \texttt{napping.csv})
  \item \texttt{year}: year the wine was produced
  \item \texttt{winery\_id}: id of the winery that produced the wine
  \item \texttt{wine}: name of the wine
  \item \texttt{alcohol}: the wine's alcohol percentage
  \item \texttt{country}: the country where the wine was produced
  \item \texttt{region}: the region where the wine was produced
  \item \texttt{price}: price of the wine in USD (collected 05/2023)
  \item \texttt{rating}: average rating of the wine (collected 05/2023)
  \item \texttt{grape}: the wine's grape composition, represented as a comma-separated list ordered in descending sequence of the percentage contribution of each grape variety to the overall blend.
\end{itemize}

\section{Data collection details} \label{sec:data_collection_details}
The annotations in WineSensed were collected through a series of wine-tasting events attended by a total of 256 non-expert wine drinkers. Most participants were between 21-25 years old, and more than half of them were from Denmark. The experiment was conducted in accordance with the ''De Videnskabsetiske Komiteer'' (e. the Danish ethics committee for science).

\begin{figure}[h!]
    \centering
    
    \begin{subfigure}{0.24\textwidth}
        \includegraphics[width=\textwidth]{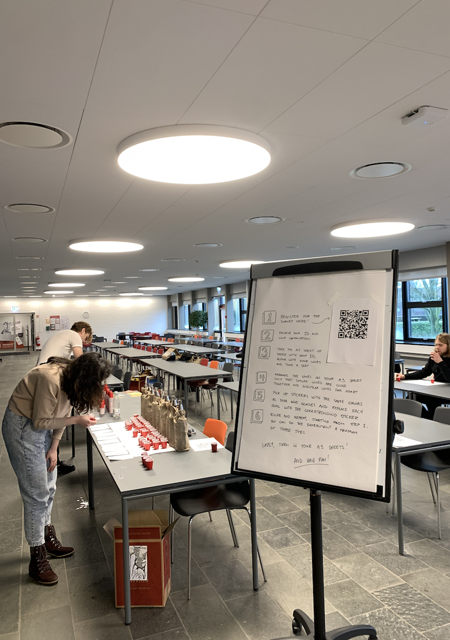}
        \caption{The participants were given instructions as to how to conduct the wine tasting, and palate cleansers were available.}
        \vspace{0.7cm}
        \label{subfig:data_collection_instructions}
    \end{subfigure}
    \hfill
    \begin{subfigure}{0.24\textwidth}
        \includegraphics[width=\textwidth]{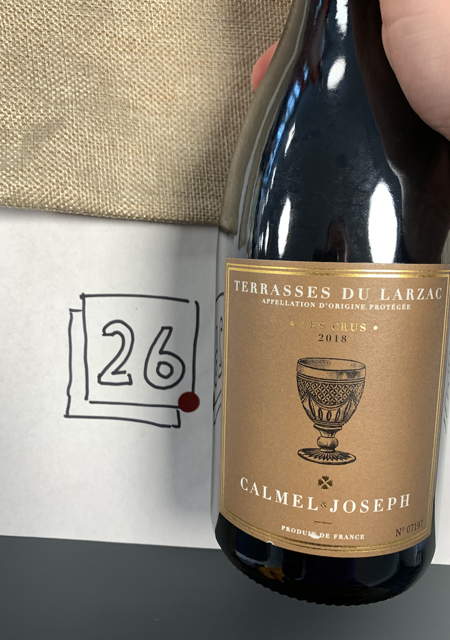}
        \caption{Each wine was labelled with a color and a number, and each participant was given a combination of wines to taste. In total, 108 different wines were used in the wine tastings.}
        \label{subfig:data_collection_labelling}
    \end{subfigure}
    \hfill
    \begin{subfigure}{0.24\textwidth}
        \includegraphics[width=\textwidth]{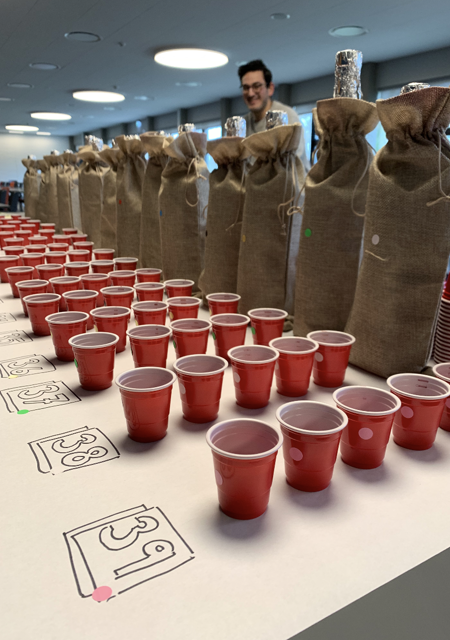}
        \caption{The portions were 15 ml each, such that each participant could taste five different wines up to three times, and still consume less than two glasses of wine.}
        \label{subfig:data_collection_portions}
    \end{subfigure}
    \hfill
    \begin{subfigure}{0.24\textwidth}
        \includegraphics[width=\textwidth]{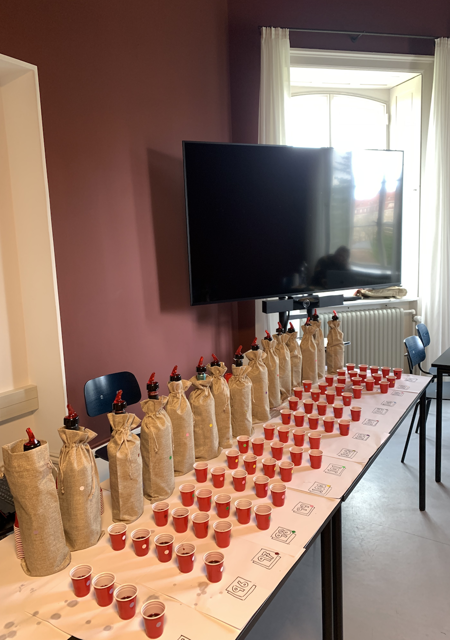}
        \caption{Anonymized wines, set up for a Napping-type data collection.}
        \vspace{1.4cm}
        \label{subfig:data_collection_setup}
    \end{subfigure}

    \caption{Data collection events. }
    \label{fig:data_collection_events}
\end{figure}

Five wines were selected at random for the participants to taste. The participants did not have access to any information regarding the individual wines. The wine was poured into non-transparent shot glasses and the labels of the wines were covered during the entire experiment. The participants were instructed to put colored stickers (representing each of the five wines) on a sheet of paper based on their taste similarity, closer meaning more similar. The participants could repeat the process up to three times, ensuring they did not consume more than 225 ml of wine. The average participant repeated the experiment two times. Figure \ref{fig:data_collection_events} provides a visual representation of the data collection events. 

\begin{figure}[h!]
    \centering
    
    \begin{subfigure}{0.24\textwidth}
        \includegraphics[width=\textwidth]{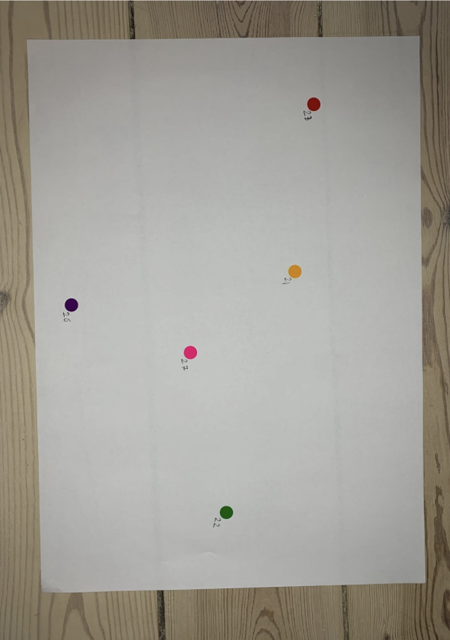}
        \vspace{1em}
        \caption{An unprocessed sample sheet with colored stickers representing the relative positions of the five different wines in the sample.}
        \label{subfig:data_collection_1}
    \end{subfigure}
    \hfill
    \begin{subfigure}{0.24\textwidth}
        \includegraphics[width=\textwidth]{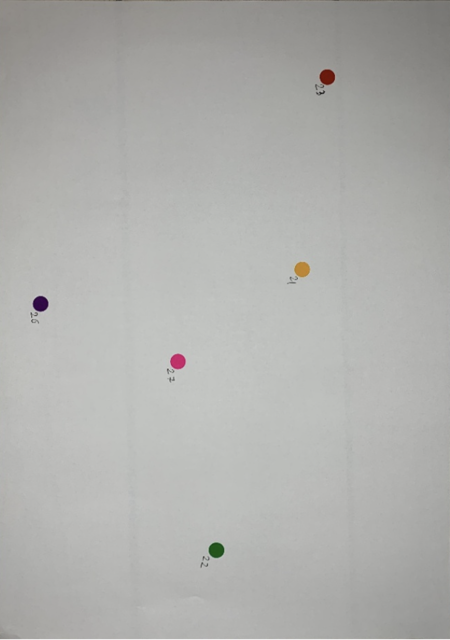}
        \vspace{1em}
        \caption{The corners of the sample sheet have been detected and used to perform perspective warping in order to correct for the angle and distance of the camera.}
        \label{subfig:data_collection_2}
    \end{subfigure}
    \hfill
    \begin{subfigure}{0.24\textwidth}
        \includegraphics[width=\textwidth]{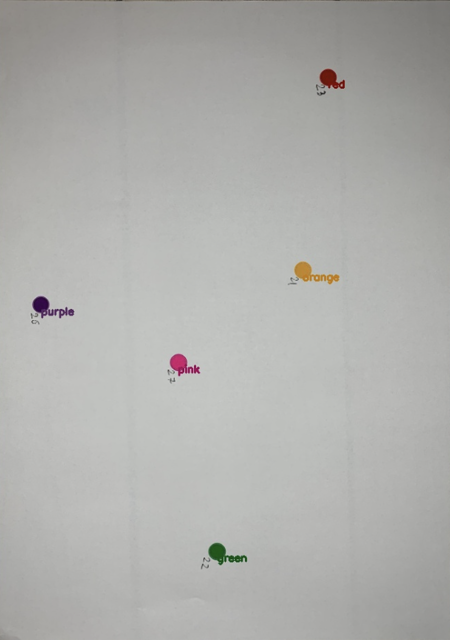}
        \vspace{1em}
        \caption{Blob detection has been performed on the perspective-corrected sample sheet and the color of each blob classified.}
        \vspace{0.4cm}
        \label{subfig:data_collection_3}
    \end{subfigure}
    \hfill
    \begin{subfigure}{0.24\textwidth}
        \includegraphics[width=\textwidth]{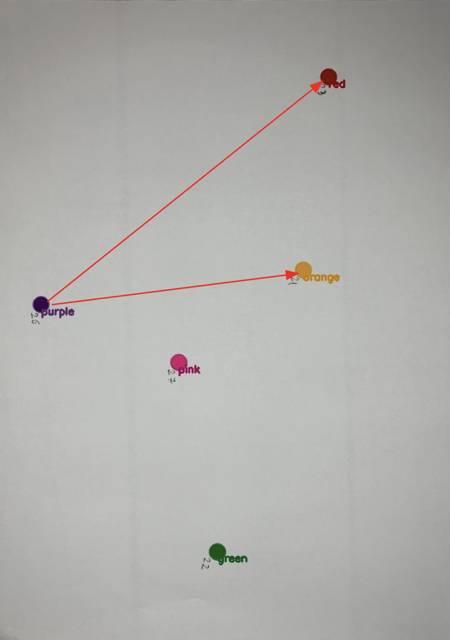}
        \vspace{1em}
        \caption{The Euclidean distance between each pair of labels is calculated.}
        \vspace{1.1cm}
        \label{subfig:data_collection_4}
    \end{subfigure}

    \caption{Napping paper analysis.}
    \label{fig:napping_papers}
\end{figure}

The participants' annotations were automatically digitized by taking a photo of each filled-out sheet. We used the Harris corner detector to find the corners of the paper and a homographic projection to obtain an aligned top-down view of the paper. The images were mapped into HSV color space and a threshold filter applied to find the different colored stickers that the participant used to represent the wines. Having identified the location, we computed the Euclidean pixel-wise distance between all pairs of points, resulting in a distance matrix of wine similarities. Figure \ref{fig:napping_papers} provides a visual representation of the procedure used to process the napping papers. 

\section{Implementation details for flavor space generation}\label{sec:training_details}
\textbf{Preprocessing. } For the image data, we resized images to a 256x256 pixel format, applied a central crop to bring the images down to 224x224 pixels. Subsequently, we converted them into a tensor format, followed by normalization using mean and standard deviation values for each color channel (RGB).

For the user reviews, we first converted the text to lowercase to maintain consistency. Then, we removed punctuation marks to minimize noise. We further eliminated stopwords using the nltk library's English stopword list since these words usually do not contribute significantly to the overall meaning of the reviews. After these preprocessing steps, the data was tokenized and reassembled into a clean text string.

The preprocessing of human-annotated data varied based on its intended use, either as a distance matrix, triplets or graph. In the first case, we calculated the Euclidean distances between each data point and arranged these distances into an $N\times N$ matrix, where $N$ is the total number of annotated wines. The matrix element $m[i][j]$ had a value of $0$ if there were no annotated distances between wines $i$ and $j$. For triplets, we constructed a list of triplets derived from the computed Euclidean distances. We generated triplets $(i, j, k)$ based on the Euclidean distances, such that $i$ is closer to $j$ than to $k$; i.e. $\|i-j\|_{2} < \|i-k\|_{2}$. In order to make a graph representation, we used the euclidean distance matrix to form a graph where the similarities between wines form the weights on the edges and the nodes represent the wines.

\textbf{Dimensionality reduction. } In our experiments, we used several dimensionality reduction methods such as NMDS, t-STE, t-SNE, PCA, and UMAP. For these methods, we prepared two embedding pipelines, one to reduce the dimensionality of machine kernel, and another to reduce the dimensionality of the human kernel.

For the human kernel, NMDS, t-STE, t-SNE and node2vec \citep{grover2016node2vec} were used, depending on the form of the input data. The NMDS method was optimized through a series of hyperparameter tunings, including number of initial positions (\texttt{n\_inits}), maximum number of iterations (\texttt{max\_iters}), and tolerance to stress convergence (\texttt{eps\_values}). These hyperparameters were evaluated using a range of values with the number of initial positions set to {5, 7, 10}, the maximum number of iterations set to {300, 400, 500, 600}, and the tolerance for stress convergence set to {1e-3, 1e-4, 1e-5}.

The optimal hyperparameters for NMDS were selected by applying 5-fold cross validation (\texttt{cross\_val\_score}) using a K-nearest neighbors classifier model (\texttt{KNeighborsClassifier}) and oversampling to handle class imbalances in the data. In NMDS, The parameter \texttt{metric} was set to \texttt{False} to handle dissimilarities missing values represented by zeroes, and \texttt{dissimilarity} to precomputed as the input data was a distance matrix. Classification improvements during grid-search were not significant.

For the machine kernel pipeline, t-SNE, PCA, and UMAP, were used with a set seed to ensure the results' reproducibility. These methods were called using their default hyperparameters in the respective libraries (see External packages).

\textbf{Pre-trained models. }
The machine kernel embeddings were obtained using a collection of pre-trained text, image, and combined image-text models. All models were obtained from the HuggingFace \citep{huggingface} library. The chosen models for the text were T5 (60.5M params), ALBERT (11.8M params), BART (406MM params), DistilBERT (67M params), FLAN-T5 (60M params), PEGASUS (568M params) and CLIP text model. For images, we chose ViT, DeiT, ResNET-50 and the CLIP image encoder. Lastly, we used CLIP for the combined image-text model. All embeddings were obtained from the models' last hidden state. 

\textbf{Combiners. }
We leveraged four methods to combine the human kernel and the machine kernel: CCA, ICP, Procrustes and SNaCK. These three methods were employed using their default hyperparameters in their respective libraries (see External packages). In the case of CCA, Procrustes and ICP, we found common experiment identifiers across the two datasets and used them to align corresponding data points from the two datasets. Once the matrices were aligned, we subsequently applied CCA, Procrustes and ICP, respectively, and generated the combined embeddings.

SNaCK follows a slightly different process as it uses triplets from the human kernel and an embedding matrix from the machine kernel. We passed the triplet list (human kernel) and scaled embeddings (machine kernel) into SNaCK, which output the combined embedding.

\textbf{External packages. } We used several external packages: \texttt{scikit-learn} (v1.2.2) \citep{scikit-learn}, for dimensionality reduction, hyperparameter optimization, classification and human-and machine kernel combination; \texttt{umap-learn} (v0.5.3) \citep{umap}, for dimensionality reduction of the machine kernel; \texttt{imblearn} (v0.10.1) \citep{imblearn}, to address the problem of imbalanced datasets; \texttt{snack} \cite{snack}, an implementation of SNaCK for human-and machine kernel combination; \texttt{icp} \citep{icp}, implementing the Iterative Closest Point algorithm for human-and machine kernel combination; \texttt{tste} \citep{tste}, an implementation of the t-Distributed Stochastic Triplet Embedding algorithm for the dimensionality reduction of human-kernel triplets; \texttt{node2vec} \citep{grover2016node2vec} for dimensionality reduction of the human kernel in graph form and \texttt{procrustes} \citep{2020SciPy-NMeth}, an implementation of the Procrustes algorithm for combining the human and machine kernel. Additionally, our project employed these Python packages: \texttt{torchmetrics} (v0.11.4) \cite{torchmetrics}, \texttt{ftfy} (v6.1.1) \citep{ftfy}, \texttt{open-clip-torch} (v2.19.0) \citep{open-clip-torch}, \texttt{transformers} (v4.28.1) \cite{transformers}, \texttt{pandas} (v2.0.1) \citep{pandas}, \texttt{nltk} (v3.8.1) \citep{nltk}, \texttt{psutil} (v5.9.5) \citep{psutil}, \texttt{urllib3} (v1.26.15) \citep{urllib3}, \texttt{matplotlib} (v3.5.1) \cite{matplotlib}, \texttt{seaborn} (v0.11.2) \citep{seaborn}, and \texttt{h5py} (v3.8.0) \citep{h5py}.

\section{Details for fine-grained flavor predictions}\label{sec:more_results}

\textbf{Implementation details. } The combination of dimensionality reduction methods, pre-trained models, and combiners described in  \ref{sec:more_results} were used to generate multiple flavor spaces (using images, text and flavor). Additionally, to compare TAR across modalities, embeddings were produced for all combinations of modalities (text, image and flavor) using the relevant methods from \ref{sec:more_results}.

The human kernel was split into a training and a testing set. We made sure that for any given triplet $(i, j, k)$ in the testing set, none of the wines $i$, $j$ or $k$ were present in the training set. The training set was processed and combined with the machine kernel using the reduction methods and combiners from \ref{sec:more_results}. The triplet agreement ratio was calculated using the level of agreement between the testing set and the triplets in the embeddings, by dividing agreements with disagreements. The triplet agreement ratio's random baseline was set at 0.5, because when comparing triplets, either (i, j, k) or (j, i, k) could be chosen, which makes the ratio 0.5/1.0, similar to a random guess.

\textbf{Results. }
All results produced in this experiment can be found in tables \ref{tab:flavor_text}, \ref{tab:fine_grained_images} and \ref{tab:fine_grained_multi}. 

\begin{table}[h]
    \centering
    \caption{\textbf{Fine-grained flavor predictions: Text encoders. } Triplet Agreement Ratio (TAR) between text encoders and human annotated flavor similarities.}
    \begin{tabular}{lll|ll} \toprule
        Machine Kernel & Human Kernel & Combiner & Modality & TAR $\uparrow$ \\ \midrule
        DistilBeRT + UMAP & & & Text only & 0.81 \\
        DistilBeRT + t-SNE & & & Text only & 0.81 \\
        DistilBeRT + UMAP & MDS & CCA & Text + flavor & \textbf{0.91} \\
        DistilBeRT + t-SNE & MDS & ICP & Text + flavor & 0.90 \\
        DistilBeRT + t-SNE & MDS & CCA & Text + flavor & 0.90 \\
        DistilBeRT + UMAP & t-STE & CCA & Text + flavor & 0.76 \\
        DistilBeRT + t-SNE & t-STE & ICP & Text + flavor & 0.78 \\
        DistilBeRT + t-SNE & t-STE & SNaCK & Text + flavor & 0.75 \\ \midrule
        T5 + UMAP & & & Text only & 0.82 \\
        T5 + t-SNE & & & Text only & 0.82 \\
        T5 + UMAP & MDS & CCA & Text + flavor & 0.89 \\
        T5 + t-SNE & MDS & ICP & Text + flavor & \textbf{0.90} \\
        T5 + t-SNE & MDS & CCA & Text + flavor & \textbf{0.90} \\
        T5 + UMAP & t-STE & CCA & Text + flavor & 0.83 \\
        T5 + t-SNE & t-STE & ICP & Text + flavor & 0.78 \\
        T5 + t-SNE & t-STE & SNaCK & Text + flavor & 0.84 \\ \midrule
        ALBERT + UMAP & & & Text only & 0.80 \\
        ALBERT + t-SNE & & & Text only & 0.81 \\
        ALBERT + UMAP & MDS & CCA & Text + flavor & 0.89 \\
        ALBERT + t-SNE & MDS & ICP & Text + flavor & \textbf{0.90} \\
        ALBERT + t-SNE & MDS & CCA & Text + flavor & \textbf{0.90} \\
        ALBERT + UMAP & t-STE & CCA & Text + flavor & 0.74 \\
        ALBERT + t-SNE & t-STE & ICP & Text + flavor & 0.78 \\
        ALBERT + t-SNE & t-STE & SNaCK & Text + flavor & 0.78 \\ \midrule
        BART + UMAP & & & Text only & 0.81 \\
        BART + t-SNE & & & Text only & 0.82 \\
        BART + UMAP & MDS & CCA & Text + flavor & 0.89 \\
        BART + t-SNE & MDS & ICP & Text + flavor & \textbf{0.90} \\
        BART + t-SNE & MDS & CCA & Text + flavor & 0.89 \\
        BART + UMAP & t-STE & CCA & Text + flavor & 0.78 \\
        BART + t-SNE & t-STE & ICP & Text + flavor & 0.79 \\
        BART + t-SNE & t-STE & SNaCK & Text + flavor  & 0.72 \\ \bottomrule
    \end{tabular}
    \label{tab:flavor_text}
\end{table}

\begin{table}[]
    \centering
    \caption{\textbf{Fine-grained flavor predictions: Image encoders. } Triplet Agreement Ratio (TAR) between image encoders and human annotated flavor similarities.}
    \begin{tabular}{lll|ll} \toprule
        Machine Kernel & Human Kernel & Combiner & Modality & TAR $\uparrow$ \\ \midrule
        ViT + UMAP & & & Image only & 0.83 \\
        ViT + t-SNE & & & Image only & 0.82 \\
        ViT + UMAP & MDS & CCA & Image + flavor & \textbf{0.90} \\
        ViT + t-SNE & MDS & ICP & Image + flavor &\textbf{0.90} \\
        ViT + t-SNE & MDS & CCA & Image + flavor &\textbf{0.90} \\
        ViT + UMAP & t-STE & CCA & Image + flavor & 0.82 \\
        ViT + t-SNE & t-STE & ICP & Image + flavor & 0.78 \\
        ViT + t-SNE & t-STE & SNaCK & Image + flavor & 0.75 \\ \midrule
        ResNET + UMAP & & & Image only & 0.82 \\
        ResNET + t-SNE & & & Image only & 0.82 \\
        ResNET + UMAP & MDS & CCA & Image + flavor & 0.89 \\
        ResNET + t-SNE & MDS & ICP & Image + flavor & \textbf{0.90} \\
        ResNET + t-SNE & MDS & CCA & Image + flavor & 0.88 \\
        ResNET + UMAP & t-STE & CCA & Image + flavor & 0.79 \\
        ResNET + t-SNE & t-STE & ICP & Image + flavor & 0.78 \\
        ResNET + t-SNE & t-STE & SNaCK & Image + flavor & 0.76 \\ \midrule
        DeiT + UMAP & & & Image only & 0.82 \\
        DeiT + t-SNE & & & Image only & 0.83 \\
        DeiT + UMAP & MDS & CCA & Image + flavor & 0.91 \\
        DeiT + t-SNE & MDS & ICP & Image + flavor & 0.90 \\
        DeiT + t-SNE & MDS & CCA & Image + flavor & \textbf{0.92} \\
        DeiT + UMAP & t-STE & CCA & Image + flavor & 0.82 \\
        DeiT + t-SNE & t-STE & ICP & Image + flavor & 0.78 \\
        DeiT + t-SNE & t-STE & SNaCK & Image + flavor & 0.86 \\ \midrule
        CLIP + UMAP & & & Image only & 0.82 \\
        CLIP + t-SNE & & & Image only & 0.82 \\
        CLIP + UMAP & MDS & CCA & Image + flavor & 0.89 \\
        CLIP + t-SNE & MDS & ICP & Image + flavor & \textbf{0.90} \\
        CLIP + t-SNE & MDS & CCA & Image + flavor & \textbf{0.90} \\
        CLIP + UMAP & t-STE & CCA & Image + flavor & 0.81 \\
        CLIP + t-SNE & t-STE & ICP & Image + flavor & 0.78 \\
        CLIP + t-SNE & t-STE & SNaCK & Image + flavor & 0.81 \\ \bottomrule
    \end{tabular}
    \label{tab:fine_grained_images}
\end{table}

\begin{table}[]
    \centering
    \caption{\textbf{Fine-grained flavor predictions: Text-Image encoder. } Triplet Agreement Ratio (TAR) between CLIP and human annotated flavor similarities.}
    \begin{tabular}{lll|ll} \toprule
        Machine Kernel & Human Kernel & Combiner & TAR Machine Kernel $\uparrow$ & TAR $\uparrow$ \\ \midrule
        CLIP + UMAP & & & Image + text & 0.82 \\
        CLIP + t-SNE & & & Image + text & 0.81 \\
        CLIP + UMAP & MDS & CCA & Image + text + flavor & \textbf{0.91} \\
        CLIP + t-SNE & MDS & ICP & Image + text + flavor & 0.90 \\
        CLIP + t-SNE & MDS & CCA & Image + text + flavor & \textbf{0.91} \\
        CLIP + UMAP & t-STE & CCA & Image + text + flavor & 0.84 \\
        CLIP + t-SNE & t-STE & ICP & Image + text + flavor & 0.78 \\
        CLIP + t-SNE & t-STE & SNaCK & Image + text + flavor & 0.79 \\ \bottomrule
    \end{tabular}
    \label{tab:fine_grained_multi}
\end{table}

\section{Details for coarse-grained flavor predictions. }\label{sec:more_classification_results}
\section{Implementation Details} We utilize a SVM classifier with parameter \texttt{class\_weight} set to balanced and and K-fold cross-validation with \texttt{n\_splits} set to 5 and \texttt{shuffle} set to True using the classifier \texttt{SVC} and the method \texttt{KFold} as well as a Multi-Layer Perceptron (\texttt{MLP}) neural network from the Scikit-Learn library \citep{scikit-learn}. Additionally we utilize \texttt{RandomOverSampler} from the imblearn library with \texttt{sampling\_strategy} set to 'not majority'. When dealing with non-numerical attributes, a \texttt{LabelEncoder} (using the default values) from Scikit-Learn \citep{scikit-learn} was used to create numerical features. The random baseline value was calculated by dividing 1 by the number of classes to predict. 

\textbf{Results. }
All results produced in this experiment can be found in tables \ref{tbl:vit}, \ref{tab:distilbert}, \ref{tab:t5}, \ref{tab:albert}, \ref{tab:bart}, \ref{tab:bart_large}, \ref{tab:resnet}, \ref{tab:deit}, \ref{tab:clip_image}, \ref{tab:pegasus}, \ref{tab:flan-t5} and \ref{tab:clip}.

\begin{table}[]
    \centering
    \small
    \caption{\textbf{ViT: } Classification results. }
    \scalebox{1.0}{
    \label{tbl:vit}
\begin{tabular}{l|l|l|l|l|c|c} \toprule
        \multicolumn{1}{c|}{} & \multicolumn{1}{c|}{} & \multicolumn{1}{c|}{} & \multicolumn{1}{c|}{} & \multicolumn{1}{c|}{} & \multicolumn{2}{c}{Acc $\uparrow$} \\ 
        Machine Kernel & Human Kernel & Combiner & Class & Modality & SVM & NN \\ 
        \midrule
        Random & & & Alc \% & & 0.17 & 0.17 \\ 
        ViT + Umap & & & Alc \% & Image & 0.03 & 0.06 \\ 
        ViT + t-SNE & & & Alc \% & Image & 0.07 & 0.06 \\ 
        ViT + Umap & t-SNE & ICP & Alc \% & Image + flavor & 0.02 & 0.04 \\ 
        ViT + t-SNE & NMDS & CCA & Alc \% & Image + flavor & \textbf{0.39} & \textbf{0.40} \\ 
        ViT + t-SNE & t-STE & SNaCK & Alc \% & Image + flavor & 0.04 & 0.13 \\
\bottomrule
    \end{tabular}
    }
\end{table}

\begin{table}[]
    \centering
    \small
    \caption{\textbf{DeiT: } Classification results. }
    \scalebox{1.0}{
    \label{tab:deit}
    \begin{tabular}{l|l|l|l|l|c|c} \toprule
        \multicolumn{1}{c|}{} & \multicolumn{1}{c|}{} & \multicolumn{1}{c|}{} & \multicolumn{1}{c|}{} & \multicolumn{1}{c|}{} & \multicolumn{2}{c}{Acc $\uparrow$} \\ 
        Machine Kernel & Human Kernel & Combiner & Class & Modality & SVM & NN \\ 
        \midrule
        Random & & & Alc \% & & 0.17 & 0.17 \\ 
        DeiT + Umap & & & Alc \% & Image & 0.17 & 0.08 \\ 
        DeiT + t-SNE & & & Alc \% & Image & 0.18 & 0.08 \\ 
        DeiT + Umap & t-SNE & ICP & Alc \% & Image + flavor & 0.13 & 0.10 \\ 
        DeiT + t-SNE & NMDS & CCA & Alc \% & Image + flavor & \textbf{0.33} & \textbf{0.38} \\ 
        DeiT + t-SNE & t-STE & SNaCK & Alc \% & Image + flavor & 0.18 & 0.17 \\
\bottomrule
    \end{tabular}
    }
\end{table}

\begin{table}[]
    \centering
    \small
    \caption{\textbf{ResNET: } Classification results. }
    \label{tab:resnet}
    \begin{tabular}{l|l|l|l|l|c|c} \toprule
        \multicolumn{1}{c|}{} & \multicolumn{1}{c|}{} & \multicolumn{1}{c|}{} & \multicolumn{1}{c|}{} & \multicolumn{1}{c|}{} & \multicolumn{2}{c}{Acc $\uparrow$} \\ 
        Machine Kernel & Human Kernel & Combiner & Class & Modality & SVM & NN \\ 
        \midrule
        Random & & & Alc \% & & 0.17 & 0.17 \\ 
        ResNET + Umap & & & Alc \% & Image & 0.07 & 0.06 \\ 
        ResNET + t-SNE & & & Alc \% & Image & 0.13 & 0.06 \\ 
        ResNET + Umap & t-SNE & ICP & Alc \% & Image + flavor & 0.08 & 0.05 \\ 
        ResNET + t-SNE & NMDS & CCA & Alc \% & Image + flavor & \textbf{0.36} & \textbf{0.37} \\ 
        ResNET + t-SNE & t-STE & SNaCK & Alc \% & Image + flavor & 0.21 & 0.21 \\
\bottomrule
    \end{tabular}
\end{table}

\begin{table}[]
    \centering
    \small
    \caption{\textbf{CLIP (image): } Classification results. }
    \label{tab:clip_image}
    \begin{tabular}{l|l|l|l|l|c|c} \toprule
        \multicolumn{1}{c|}{} & \multicolumn{1}{c|}{} & \multicolumn{1}{c|}{} & \multicolumn{1}{c|}{} & \multicolumn{1}{c|}{} & \multicolumn{2}{c}{Acc $\uparrow$} \\ 
        Machine Kernel & Human Kernel & Combiner & Class & Modality & SVM & NN \\ 
        \midrule
        Random & & & Alc \% & & 0.17 & 0.17 \\ 
        CLIP (image) + Umap & & & Alc \% & Image & 0.10 & 0.26 \\ 
        CLIP (image) + t-SNE & & & Alc \% & Image & 0.22 & 0.23 \\ 
        CLIP (image) + Umap & t-SNE & ICP & Alc \% & Image + flavor & 0.13 & 0.14 \\ 
        CLIP (image) + t-SNE & NMDS & CCA & Alc \% & Image + flavor & \textbf{0.38} & \textbf{0.39} \\ 
        CLIP (image) + t-SNE & t-STE & SNaCK & Alc \% & Image + flavor & 0.14 & 0.15 \\
\bottomrule
    \end{tabular}
\end{table}

\begin{table}[]
    \centering
    \small
    \caption{\textbf{PEGASUS: } Classification results. }
    \label{tab:pegasus}
    \begin{tabular}{l|l|l|l|l|c|c} \toprule
        \multicolumn{1}{c|}{} & \multicolumn{1}{c|}{} & \multicolumn{1}{c|}{} & \multicolumn{1}{c|}{} & \multicolumn{1}{c|}{} & \multicolumn{2}{c}{Acc $\uparrow$} \\ 
        Machine Kernel & Human Kernel & Combiner & Class & Modality & SVM & NN \\ 
        \midrule
        Random & & & Alc \% & & 0.17 & 0.17 \\ 
        PEGASUS + Umap & & & Alc \% & Text & 0.25 & 0.15 \\ 
        PEGASUS + t-SNE & & & Alc \% & Text & 0.23 & 0.19 \\ 
        PEGASUS + Umap & t-SNE & ICP & Alc \% & Text + flavor & 0.21 & 0.21 \\ 
        PEGASUS + t-SNE & NMDS & CCA & Alc \% & Text + flavor & \textbf{0.41} & \textbf{0.42} \\ 
        PEGASUS + t-SNE & t-STE & SNaCK & Alc \% & Text + flavor & 0.28 & 0.29 \\
\bottomrule
    \end{tabular}
\end{table}

\begin{table}[]
    \centering
    \small
    \caption{\textbf{BART (large): } Classification results. }
    \label{tab:bart_large}
    \begin{tabular}{l|l|l|l|l|c|c} \toprule
        \multicolumn{1}{c|}{} & \multicolumn{1}{c|}{} & \multicolumn{1}{c|}{} & \multicolumn{1}{c|}{} & \multicolumn{1}{c|}{} & \multicolumn{2}{c}{Acc $\uparrow$} \\ 
        Machine Kernel & Human Kernel & Combiner & Class & Modality & SVM & NN \\ 
        \midrule
        Random & & & Alc \% & & 0.17 & 0.17 \\ 
        BART (large) + Umap & & & Alc \% & Text & 0.21 & 0.30 \\ 
        BART (large) + t-SNE & & & Alc \% & Text & 0.13 & 0.28 \\ 
        BART (large) + Umap & t-SNE & ICP & Alc \% & Text + flavor & 0.15 & 0.08 \\ 
        BART (large) + t-SNE & NMDS & CCA & Alc \% & Text + flavor & \textbf{0.40} & \textbf{0.39} \\ 
        BART (large) + t-SNE & t-STE & SNaCK & Alc \% & Text + flavor & 0.25 & 0.17 \\
\bottomrule
    \end{tabular}
\end{table}

\begin{table}[]
    \centering
    \small
    \caption{\textbf{FLAN-T5: } Classification results. }
    \label{tab:flan-t5}
    \begin{tabular}{l|l|l|l|l|c|c} \toprule
        \multicolumn{1}{c|}{} & \multicolumn{1}{c|}{} & \multicolumn{1}{c|}{} & \multicolumn{1}{c|}{} & \multicolumn{1}{c|}{} & \multicolumn{2}{c}{Acc $\uparrow$} \\ 
        Machine Kernel & Human Kernel & Combiner & Class & Modality & SVM & NN \\ 
        \midrule
        Random & & & Alc \% & & 0.17 & 0.17 \\ 
        FLAN-T5 + Umap & & & Alc \% & Text & 0.10 & 0.13 \\ 
        FLAN-T5 + t-SNE & & & Alc \% & Text & \textbf{0.47} & 0.40 \\ 
        FLAN-T5 + Umap & t-SNE & ICP & Alc \% & Text + flavor & 0.22 & 0.27 \\ 
        FLAN-T5 + t-SNE & NMDS & CCA & Alc \% & Text + flavor & 0.46 & \textbf{0.51} \\ 
        FLAN-T5 + t-SNE & t-STE & SNaCK & Alc \% & Text + flavor & 0.39 & 0.41 \\
\bottomrule
    \end{tabular}
\end{table}

\begin{table}[]
    \centering
    \small
    \caption{\textbf{DistilBERT: } Classification results. }
    \label{tab:distilbert}
    \begin{tabular}{l|l|l|l|l|c|c} \toprule
        \multicolumn{1}{c|}{} & \multicolumn{1}{c|}{} & \multicolumn{1}{c|}{} & \multicolumn{1}{c|}{} & \multicolumn{1}{c|}{} & \multicolumn{2}{c}{Acc $\uparrow$} \\ 
        Machine Kernel & Human Kernel & Combiner & Class & Modality & SVM & NN \\ 
        \midrule
        Random & & & Alc \% & & 0.17 & 0.17 \\ 
        DistilBERT + Umap & & & Alc \% & Text & 0.27 & 0.31 \\ 
        DistilBERT + t-SNE & & & Alc \% & Text & 0.25 & 0.24 \\ 
        DistilBERT + Umap & t-SNE & ICP & Alc \% & Text + flavor & 0.24 & 0.25 \\ 
        DistilBERT + t-SNE & NMDS & CCA & Alc \% & Text + flavor & \textbf{0.46} & \textbf{0.39} \\ 
        DistilBERT + t-SNE & t-STE & SNaCK & Alc \% & Text + flavor & 0.33 & 0.29 \\
\bottomrule
    \end{tabular}
\end{table}

\begin{table}[]
    \centering
    \small
    \caption{\textbf{T5: } Classification results. }
    \label{tab:t5}
    \begin{tabular}{l|l|l|l|l|c|c} \toprule
        \multicolumn{1}{c|}{} & \multicolumn{1}{c|}{} & \multicolumn{1}{c|}{} & \multicolumn{1}{c|}{} & \multicolumn{1}{c|}{} & \multicolumn{2}{c}{Acc $\uparrow$} \\ 
        Machine Kernel & Human Kernel & Combiner & Class & Modality & SVM & NN \\ 
        \midrule
        Random & & & Alc \% & & 0.17 & 0.17 \\ 
        T5 + Umap & & & Alc \% & Text & 0.24 & 0.14 \\ 
        T5 + t-SNE & & & Alc \% & Text & 0.30 & 0.34 \\ 
        T5 + Umap & t-SNE & ICP & Alc \% & Text + flavor & 0.19 & 0.17 \\ 
        T5 + t-SNE & NMDS & CCA & Alc \% & Text + flavor & \textbf{0.47} & \textbf{0.46} \\ 
        T5 + t-SNE & t-STE & SNaCK & Alc \% & Text + flavor & 0.31 & 0.32 \\
\bottomrule
    \end{tabular}
\end{table}

\begin{table}[]
    \centering
    \small
    \caption{\textbf{ALBERT: } Classification results. }
    \label{tab:albert}
    \begin{tabular}{l|l|l|l|l|c|c} \toprule
        \multicolumn{1}{c|}{} & \multicolumn{1}{c|}{} & \multicolumn{1}{c|}{} & \multicolumn{1}{c|}{} & \multicolumn{1}{c|}{} & \multicolumn{2}{c}{Acc $\uparrow$} \\ 
        Machine Kernel & Human Kernel & Combiner & Class & Modality & SVM & NN \\ 
        \midrule
        Random & & & Alc \% & & 0.17 & 0.17 \\ 
        ALBERT + Umap & & & Alc \% & Text & 0.24 & 0.17 \\ 
        ALBERT + t-SNE & & & Alc \% & Text & 0.23 & 0.28 \\ 
        ALBERT + Umap & t-SNE & ICP & Alc \% & Text + flavor & 0.10 & 0.25 \\ 
        ALBERT + t-SNE & NMDS & CCA & Alc \% & Text + flavor & \textbf{0.46} & \textbf{0.51} \\ 
        ALBERT + t-SNE & t-STE & SNaCK & Alc \% & Text + flavor & 0.20 & 0.15 \\
\bottomrule
    \end{tabular}
\end{table}

\begin{table}[]
    \centering
    \small
    \caption{\textbf{BART (small): } Classification results. }
    \label{tab:bart}
    \begin{tabular}{l|l|l|l|l|c|c} \toprule
        \multicolumn{1}{c|}{} & \multicolumn{1}{c|}{} & \multicolumn{1}{c|}{} & \multicolumn{1}{c|}{} & \multicolumn{1}{c|}{} & \multicolumn{2}{c}{Acc $\uparrow$} \\ 
        Machine Kernel & Human Kernel & Combiner & Class & Modality & SVM & NN \\ 
        \midrule
        Random & & & Alc \% & & 0.17 & 0.17 \\ 
        BART + Umap & & & Alc \% & Text & 0.14 & 0.15 \\ 
        BART + t-SNE & & & Alc \% & Text & 0.27 & 0.33 \\ 
        BART + Umap & t-SNE & ICP & Alc \% & Text + flavor & 0.30 & 0.19 \\ 
        BART + t-SNE & NMDS & CCA & Alc \% & Text + flavor & \textbf{0.47} & \textbf{0.44} \\ 
        BART + t-SNE & t-STE & SNaCK & Alc \% & Text + flavor & 0.21 & 0.14 \\
\bottomrule
    \end{tabular}
\end{table}

\begin{table}[]
    \centering
    \small
    \caption{\textbf{CLIP (text): } Classification results. }
    \label{tab:clip_text}
    \begin{tabular}{l|l|l|l|l|c|c} \toprule
        \multicolumn{1}{c|}{} & \multicolumn{1}{c|}{} & \multicolumn{1}{c|}{} & \multicolumn{1}{c|}{} & \multicolumn{1}{c|}{} & \multicolumn{2}{c}{Acc $\uparrow$} \\ 
        Machine Kernel & Human Kernel & Combiner & Class & Modality & SVM & NN \\ 
        \midrule
        Random & & & Alc \% & & 0.17 & 0.17 \\ 
        CLIP (text) + Umap & & & Alc \% & Text & 0.16 & 0.06 \\ 
        CLIP (text) + t-SNE & & & Alc \% & Text & 0.23 & 0.24 \\ 
        CLIP (text) + Umap & t-SNE & ICP & Alc \% & Text + flavor & 0.15 & 0.10 \\ 
        CLIP (text) + t-SNE & NMDS & CCA & Alc \% & Text + flavor & \textbf{0.51} & \textbf{0.45} \\ 
        CLIP (text) + t-SNE & t-STE & SNaCK & Alc \% & Text + flavor & 0.19 & 0.20 \\
\bottomrule
    \end{tabular}
\end{table}

\begin{table}[]
    \centering
    \small
    \caption{\textbf{CLIP (image and text): } Classification results. }
    \label{tab:clip}
    \begin{tabular}{l|l|l|l|l|c|c} \toprule
        \multicolumn{1}{c|}{} & \multicolumn{1}{c|}{} & \multicolumn{1}{c|}{} & \multicolumn{1}{c|}{} & \multicolumn{1}{c|}{} & \multicolumn{2}{c}{Acc $\uparrow$} \\ 
        Machine Kernel & Human Kernel & Combiner & Class & Modality & SVM & NN \\ 
        \midrule
        Random & & & Alc \% & & 0.17 & 0.17 \\ 
        CLIP + Umap & & & Alc \% & Image + text & 0.27 & 0.06 \\ 
        CLIP + t-SNE & & & Alc \% & Image + text & 0.34 & 0.36 \\ 
        CLIP + Umap & t-SNE & ICP & Alc \% & Image + text + flavor & 0.05 & 0.13 \\ 
        CLIP + t-SNE & NMDS & CCA & Alc \% & Image + text + flavor & \textbf{0.54} & \textbf{0.53} \\ 
        CLIP + t-SNE & t-STE & SNaCK & Alc \% & Image + text + flavor & 0.24 & 0.24 \\
\bottomrule
    \end{tabular}
\end{table}

\newpage
\section{Ethical approval}\label{sec:ethics}
\begin{figure}[h]
    \centering
    \begin{subfigure}[b]{0.3\textwidth}
        \includegraphics[width=\textwidth,frame]{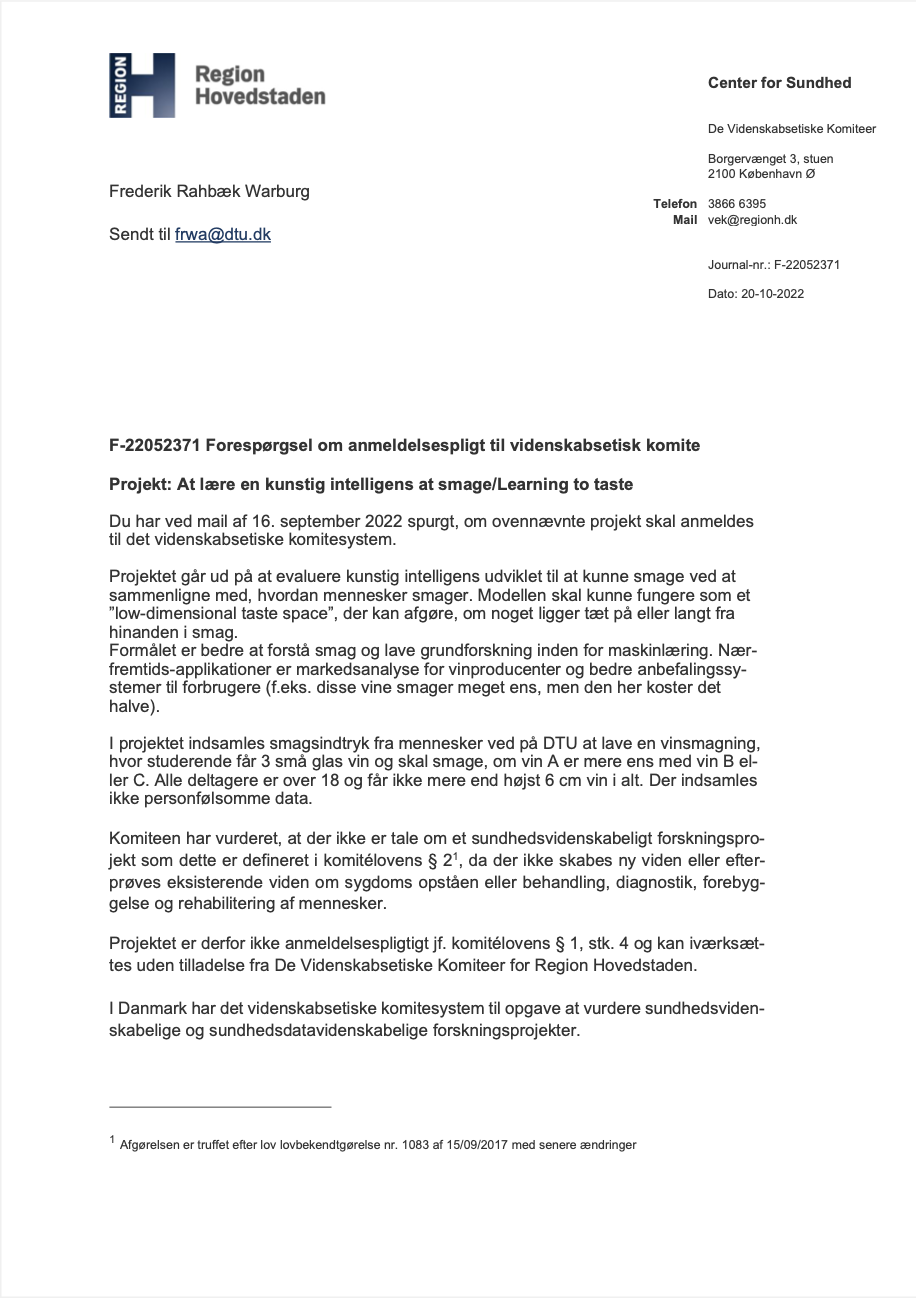}
        \caption{Page 1}
        \label{fig:subfig1}
    \end{subfigure}
    \begin{subfigure}[b]{0.3\textwidth}
        \includegraphics[width=\textwidth,frame]{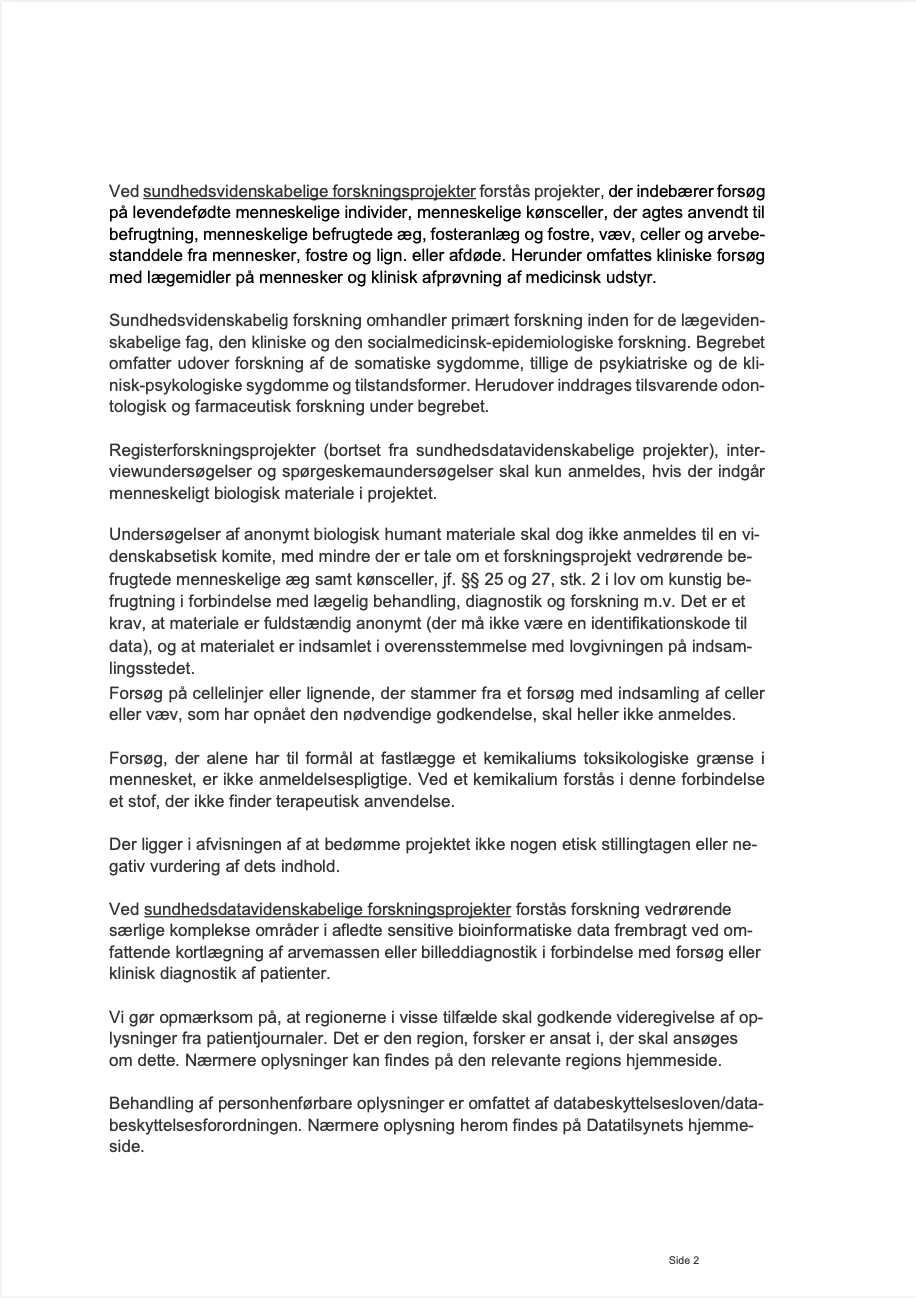}
        \caption{Page 2}
        \label{fig:subfig2}
    \end{subfigure}
    \begin{subfigure}[b]{0.3\textwidth}
        \includegraphics[width=\textwidth,frame]{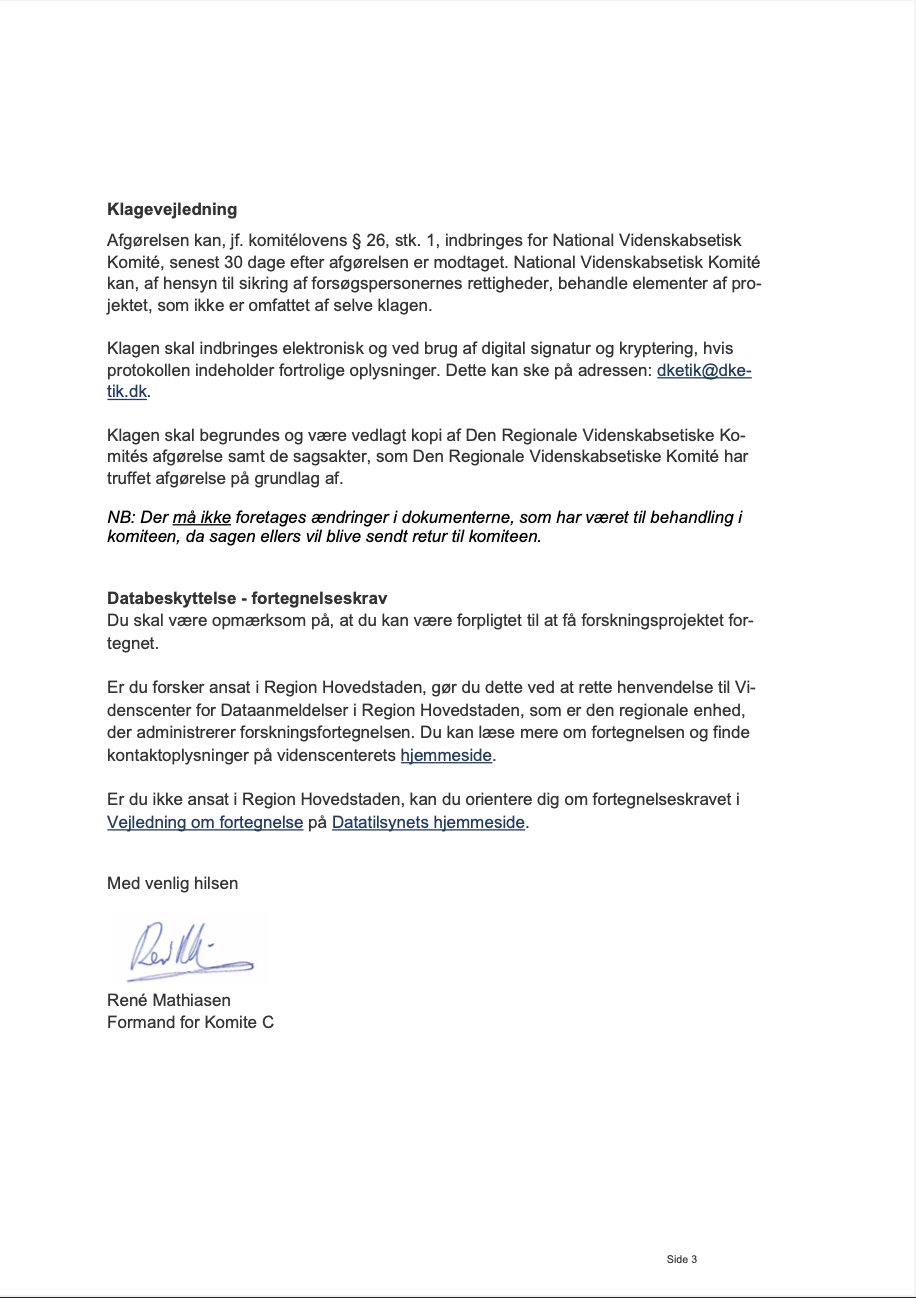}
        \caption{Page 3}
        \label{fig:subfig3}
    \end{subfigure}
    \caption{Ethical Approval (in Danish). }
    \label{fig:ethical_approval}
\end{figure}

The original ethical approval is shown in Figure \ref{fig:ethical_approval} English translation of the ethical approval can be found in section \ref{sec:english-translation}. 

\subsection{English translation}\label{sec:english-translation}
\textbf{F-22052371 Inquiry Regarding Reporting Obligations to the Ethical Scientific Committee}

\textbf{Project: Learning to Taste}

You have asked via email on September 16, 2022, if the above-mentioned project must be reported to the Ethical Scientific Committee. The project involves evaluating an artificial intelligence developed to mimic the human ability to taste, comparing it with the way humans experience flavors. The model should function as a "low-dimensional taste space", which can determine whether something is close to or far from each other in terms of taste.

The aim is to better understand taste and conduct basic research in machine learning. Near-future applications include market analysis for wine producers and improved recommendation systems for consumers (e.g., these wines taste very similar, but this one costs half as much).

In the project, taste impressions from humans are collected by conducting a wine tasting at DTU, where students are given three small glasses of wine to taste whether wine A is more similar to wine B or C. All participants are over 18 and receive no more than a maximum of 6 cl of wine in total. No sensitive personal data is collected.

The committee has assessed that this is not a health science research project as defined in the committee law's section 21, as it does not create new knowledge or test existing knowledge about disease onset or treatment, diagnostics, prevention, and rehabilitation of humans. 

Therefore, the project is not subject to reporting according to the committee law's section 1, paragraph 4 and can be implemented without permission from the Ethical Scientific Committees for the Capital Region of Denmark.

In Denmark, the task of the Ethical Scientific Committee system is to assess health science and health data science research projects.

Health science research projects refer to experiments involving live-born human individuals, human gametes intended for fertilization, human fertilized eggs, embryonic and fetal tissues, cells, and hereditary components from humans, fetuses, and the like, or deceased individuals. This includes clinical trials with drugs on humans and clinical testing of medical equipment. 

Health science research primarily covers research in the field of medical science, clinical, and social-medical epidemiological research. In addition to research on somatic diseases, the term also encompasses psychiatric and clinical-psychological diseases and conditions. Correspondingly, dental and pharmaceutical research are included under the term. 

Registered research projects (except for health data science projects), interviews, and questionnaire surveys only need to be reported if human biological material is included in the project. However, investigations of anonymous biological human material do not need to be reported to an ethical scientific committee unless the research project relates to fertilized human eggs and sex cells, cf. sections 25 and 27, paragraph 2 in the Act on Artificial Fertilization in connection with medical treatment, diagnosis, and research. It is a requirement that the material is completely anonymous (there must not be an identification code for data), and that the material is collected in accordance with the law at the collection site.

Experiments on cell lines or similar originating from an experiment collecting cells or tissue, which has received the necessary approval, also do not need to be reported. Experiments that aim solely to determine a chemical's toxicological limit in humans do not need to be reported. In this context, a chemical is understood to mean a substance that does not find therapeutic use.

The rejection to review the project does not imply an ethical stance or negative assessment of its content.

Health data science research projects refer to research concerning particular complex areas of derived sensitive bio-information data produced by comprehensive mapping of the genetic mass or imaging diagnostics in connection with experiments or clinical diagnostics of patients.

We note that in certain cases, the regions must approve the disclosure of information from patient records. The region in which the researcher is employed must be applied to for this. More information can be found on the relevant region's website.

The processing of identifiable personal information is subject to the Data Protection Act/Data Protection Regulation. More information about this can be found on the Danish Data Protection Agency's website.

According to section 26, paragraph 1 of the Committee Act, the decision can be appealed to the National Ethical Scientific Committee no later than 30 days after the decision has been received. The National Ethical Scientific Committee may, for the sake of safeguarding the rights of the test subjects, handle aspects of the project not covered by the appeal itself.

Appeals must be filed electronically and using a digital signature and encryption if the protocol contains confidential information. This can be done at the address: dketik@dke-tik.dk.

The appeal must be justified and accompanied by a copy of the decision of the Regional Ethical Scientific Committee and the case documents on which the Regional Ethical Scientific Committee has made its decision. 

\textbf{Note:} No changes should be made to the documents that have been reviewed by the committee, otherwise, the case will be returned to the committee.

\textbf{Data Protection - Registry Requirement}

Please note that you may be required to register the research project. 

If you are a researcher employed in the Capital Region, you do this by contacting the Knowledge Center for Data Reviews in the Capital Region, which is the regional unit that administers the research registry. You can read more about the registry and find contact information on the knowledge center's website. 

If you are not employed in the Capital Region, you can learn about the registry requirement in the Guide to the Registry on the Data Inspectorate's website. 

\begin{flushright}
Best regards,\\
René Mathiasen\\
Chairman of Committee C
\end{flushright}

\section{Datasheet} \label{sec:datasetdoc}

\subsection{Motivation}
\textbf{For what purpose was the dataset created?}
% Was there a specific task in mind? Was there a specific gap that needed to be filled? Please provide a description.

\textcolor{blue}{\textbf{Answer: }} The dataset was created to bridge the gap between food science and machine learning communities and introduce flavor as a modality in multimodal models. 

\textbf{Who created the dataset (e.g., which team, research group) and on behalf of which entity (e.g., company, institution, organization)?}

\textcolor{blue}{\textbf{Answer: }} Eight researchers at the Technical University of Denmark, University of Copenhagen, Vivino and California Institute of Technology have created the dataset: Thoranna Bender, Simon Moe Søresen, Alireza Kashani, Kristjan Eldjarn Hjorleifsson, Grethe Hyldig, Søren Hauberg and Frederik Warburg. 

\textbf{Who funded the creation of the dataset?}
% If there is an associated grant, please provide the name of the grantor and the grant name and number.

\textcolor{blue}{\textbf{Answer: }} The dataset is funded in part by The Danish Data Science Academy (DDSA) and the Pioneer Centre for AI (DNRF grant number P1). 

\textbf{Any other comments?}

\textcolor{blue}{\textbf{Answer: }} No. 

\subsection{Composition}
% Most of these questions are intended to provide dataset consumers with the information they need to make informed decisions about using the dataset for specific tasks. The answers to some of these questions reveal information about compliance with the EU’s General Data Protection Regulation (GDPR) or comparable regulations in other jurisdictions.

\textbf{What do the instances that comprise the dataset represent (e.g., documents, photos, people, countries)?}
% Are there multiple types of instances (e.g., movies, users, and ratings; people and interactions between them; nodes and edges)? Please provide a description.

\textcolor{blue}{\textbf{Answer: }} Each instance is an image of a wine bottle, a review about the wine, position of the wines on napping papers and attributes (grape, country, region, alcohol \%, price and rating). 

\textbf{How many instances are there in total (of each type, if appropriate)?}

\textcolor{blue}{\textbf{Answer: }} 897k images, 824k reviews of 350k vintages, around 5\% of which are also associated with year, region, rating, alcohol percentage, and grape composition. In addition there are over 5k annotated pairwise flavor distances for 108 of the wines.

\textbf{Does the dataset contain all possible instances or is it a sample (not necessarily random) of instances from a larger set?}
% If the dataset is a sample, then what is the larger set? Is the sample representative of the larger set (e.g., geographic coverage)? If so, please describe how this representativeness was validated/verified. If it is not representative of the larger set, please describe why not (e.g., to cover a more diverse range of instances, because instances were withheld or unavailable).

\textcolor{blue}{\textbf{Answer: }} The provided images, reviews and attributes are sampled from Vivino's database. The provided flavor annotations are provided in full for the 108 wines they exist for. 

\textbf{What data does each instance consist of?}
% “Raw” data (e.g., unprocessed text or images) or features? In either case, please provide a description.

\textcolor{blue}{\textbf{Answer: }} The images are .jpg files, the reviews are unprocessed text, the attributes are either numerical or categorical fields and the flavor annotations are numerical x-axis and y-axix position annotations. 

\textbf{Is there a label or target associated with each instance?}
% If so, please provide a description.

\textcolor{blue}{\textbf{Answer: }} No, but attributes can be used as targets as shown in section \label{sec:more_classification_results}. 

\textbf{Is any information missing from individual instances?}
% f so, please provide a description, explaining why this information is missing (e.g., because it was unavailable). This does not include intentionally removed information, but might include, e.g., redacted text.

\textcolor{blue}{\textbf{Answer: }} Yes, the attributes are available for approximately 5\% of the dataset and the flavor annotations are available for 108 vintages in the dataset. 

\textbf{Are relationships between individual instances made explicit (e.g., users’ movie ratings, social network links)?}
% f so, please describe how these relationships are made explicit.

\textcolor{blue}{\textbf{Answer: }} Yes, participant ID's are mappable to flavor annotations by using the values in the session\_round\_name, experiment\_round and experiment\_no fields in participants.csv and napping.csv. 

\textbf{Are there recommended data splits (e.g., training, development/validation, testing)?}
% If so, please provide a description of these splits, explaining the rationale behind them.

\textcolor{blue}{\textbf{Answer: }} No. 

\textbf{Are there any errors, sources of noise, or redundancies in the dataset?}
% If so, please provide a description.

\textcolor{blue}{\textbf{Answer: }} No. 

\textbf{Is the dataset self-contained, or does it link to or otherwise rely on external resources (e.g., websites, tweets, other datasets)?}
% If it links to or relies on external resources, a) are there guarantees that they will exist, and remain constant, over time; b) are there official archival versions of the complete dataset (i.e., including the external resources as they existed at the time the dataset was created); c) are there any restrictions (e.g., licenses, fees) associated with any of the external resources that might apply to a future user? Please provide descriptions of all external resources and any restrictions associated with them, as well as links or other access points, as appropriate.

\textcolor{blue}{\textbf{Answer: }} The data is self-contained. 

\textbf{Does the dataset contain data that might be considered confidential (e.g., data that is protected by legal privilege or by doctor-patient confidentiality, data that includes the content of individuals’ non-public communications)?}
% If so, please provide a description.

\textcolor{blue}{\textbf{Answer: }} No. 

\textbf{Does the dataset contain data that, if viewed directly, might be offensive, insulting, threatening, or might otherwise cause anxiety?}
% If so, please describe why.

\textcolor{blue}{\textbf{Answer: }} No. 

\textbf{Does the dataset relate to people?}
% If not, you may skip the remaining questions in this section.

\textcolor{blue}{\textbf{Answer: }} Yes, but indirectly. Reviews, images and flavor annotations could provide some indirect information about the people annotating them (such as language used in reviews or background in images) but no attributes containing specific information about the people (such as gender, country, age etc.) exists in the dataset. 

\textbf{Does the dataset identify any subpopulations (e.g., by age, gender)?}
% If so, please describe how these subpopulations are identified and provide a description of their respective distributions within the dataset.

\textcolor{blue}{\textbf{Answer: }} No. 

\textbf{Is it possible to identify individuals (i.e., one or more natural persons), either directly or indirectly (i.e., in combination with other data) from the dataset?}
% If so, please describe how.

\textcolor{blue}{\textbf{Answer: }} No. 

\textbf{Does the dataset contain data that might be considered sensitive in any way (e.g., data that reveals racial or ethnic origins, sexual orientations, religious beliefs, political opinions or union memberships, or locations; financial or health data; biometric or genetic data; forms of government identification, such as social security numbers; criminal history)?}
% If so, please provide a description.

\textcolor{blue}{\textbf{Answer: }} No. 

\textbf{Any other comments?}

\textcolor{blue}{\textbf{Answer: }} No. 

\subsection{Collection process}

% [T]he answers to questions here may provide information that allow others to reconstruct the dataset without access to it.

\textbf{How was the data associated with each instance acquired?}
% Was the data directly observable (e.g., raw text, movie ratings), reported by subjects (e.g., survey responses), or indirectly inferred/derived from other data (e.g., part-of-speech tags, model-based guesses for age or language)? If data was reported by subjects or indirectly inferred/derived from other data, was the data validated/verified? If so, please describe how.

\textcolor{blue}{\textbf{Answer: }} The flavor data was reported by subjects using the Napping method. The images, reviews and attributes were fetched from the Vivino platform. The flavor data was verified by a human manually checking the correctness of the algorithms annotating the napping papers. The attributes have been verified by a human to correctly represent the information about individual vintages available on the Vivino platform. 

\textbf{What mechanisms or procedures were used to collect the data (e.g., hardware apparatus or sensor, manual human curation, software program, software API)?}
% How were these mechanisms or procedures validated?

\textcolor{blue}{\textbf{Answer: }} Manual human curation and information fetched from Vivino's databases. 

\textbf{If the dataset is a sample from a larger set, what was the sampling strategy (e.g., deterministic, probabilistic with specific sampling probabilities)?}

\textcolor{blue}{\textbf{Answer: }} Not applicable.

\textbf{Who was involved in the data collection process (e.g., students, crowdworkers, contractors) and how were they compensated (e.g., how much were crowdworkers paid)?}

\textcolor{blue}{\textbf{Answer: }} Crowd-workers that volunteered their time annotated the flavor distances. Alireza Kashani provided the image- and review data on behalf of Vivino. Attributes for the wines were collected from the Vivino platform. 

\textbf{Over what timeframe was the data collected?}
% Does this timeframe match the creation timeframe of the data associated with the instances (e.g. recent crawl of old news articles)? If not, please describe the timeframe in which the data associated with the instances was created.

\textcolor{blue}{\textbf{Answer: }} The data was collected over the timeframe of June 2022 to May 2023.

\textbf{Were any ethical review processes conducted (e.g., by an institutional review board)?}
% If so, please provide a description of these review processes, including the outcomes, as well as a link or other access point to any supporting documentation.

\textcolor{blue}{\textbf{Answer: }} Yes, the ethical approval is provided in \ref{sec:ethics}. 

\textbf{Does the dataset relate to people?}
% If not, you may skip the remainder of the questions in this section.

\textcolor{blue}{\textbf{Answer: }} Yes. 

\textbf{Did you collect the data from the individuals in question directly, or obtain it via third parties or other sources (e.g., websites)?}

\textcolor{blue}{\textbf{Answer: }} Obtained from the individuals directly.

\textbf{Were the individuals in question notified about the data collection?}
% If so, please describe (or show with screenshots or other information) how notice was provided, and provide a link or other access point to, or otherwise reproduce, the exact language of the notification itself.

\textcolor{blue}{\textbf{Answer: }} Yes. 

\textbf{Did the individuals in question consent to the collection and use of their data?}
% If so, please describe (or show with screenshots or other information) how consent was requested and provided, and provide a link or other access point to, or otherwise reproduce, the exact language to which the individuals consented.

\textcolor{blue}{\textbf{Answer: }} Yes. 

\textbf{If consent was obtained, were the consenting individuals provided with a mechanism to revoke their consent in the future or for certain uses?}
% If so, please provide a description, as well as a link or other access point to the mechanism (if appropriate).

\textcolor{blue}{\textbf{Answer: }} No, this was not considered necessary, as the data can not be traced back to individuals. 

\textbf{Has an analysis of the potential impact of the dataset and its use on data subjects (e.g., a data protection impact analysis) been conducted?}
% If so, please provide a description of this analysis, including the outcomes, as well as a link or other access point to any supporting documentation.

\textcolor{blue}{\textbf{Answer: }} No. 

\textbf{Any other comments?}

\textcolor{blue}{\textbf{Answer: }} No. 

\subsection{Preprocessing/cleaning/labeling}
% The questions in this section are intended to provide dataset consumers with the information they need to determine whether the “raw” data has been processed in ways that are compatible with their chosen tasks. For example, text that has been converted into a “bag-of-words” is not suitable for tasks involving word order.

\textbf{Was any preprocessing/cleaning/labeling of the data done (e.g., discretization or bucketing, tokenization, part-of-speech tagging, SIFT feature extraction, removal of instances, processing of missing values)?}
% If so, please provide a description. If not, you may skip the remainder of the questions in this section.

\textcolor{blue}{\textbf{Answer: }} Yes, flavor annotation sample sheets from crowd-workers were digitized, by using the Harris corner detector~\citep{harris1988combined} to find the corners of the paper and a homographic projection to obtain an aligned top-down view of the paper. The images were mapped into HSV color space and a threshold filter applied to find the different colored stickers that the participant used to represent the wines. Having identified the location, we provide the Euclidean pixel-wise distance between all pairs of points in the dataset.

\textbf{Was the “raw” data saved in addition to the preprocessed/cleaned/labeled data (e.g., to support unanticipated future uses)?}
% If so, please provide a link or other access point to the “raw” data.

\textcolor{blue}{\textbf{Answer: }} No, the sample sheets themselves were deemed to contain no information in addition to the pairwise distances provided.

\textbf{Is the software used to preprocess/clean/label the instances available?}
% If so, please provide a link or other access point.

\textcolor{blue}{\textbf{Answer: }} Yes, the preprocessing software is available at \texttt{https://github.com/thoranna/learning\_to\_taste}.

\textbf{Any other comments?}

\textcolor{blue}{\textbf{Answer: }} No.

\subsection{Uses}

% These questions are intended to encourage dataset creators to reflect on the tasks for which the dataset should and should not be used. By explicitly highlighting these tasks, dataset creators can help dataset consumers to make informed decisions, thereby avoiding potential risks or harms.

\textbf{Has the dataset been used for any tasks already?}
% If so, please provide a description.

\textcolor{blue}{\textbf{Answer: }} Yes, the dataset has been used to classify different wines according to the attributes provided in the dataset.

\textbf{Is there a repository that links to any or all papers or systems that use the dataset?}
% If so, please provide a link or other access point.

\textcolor{blue}{\textbf{Answer: }} Yes, the analysis performed is available at \texttt{https://github.com/thoranna/learning\_to\_taste}.

\textbf{What (other) tasks could the dataset be used for?}

\textcolor{blue}{\textbf{Answer: }} The dataset could be used for analyzing how similar different peoples' sense of taste is. It could also be used to identify wines that taste similar, but are available at different price points.

\textbf{Is there anything about the composition of the dataset or the way it was collected and preprocessed/cleaned/labeled that might impact future uses?}
% For example, is there anything that a future user might need to know to avoid uses that could result in unfair treatment of individuals or groups (e.g., stereotyping, quality of service issues) or other undesirable harms (e.g., financial harms, legal risks) If so, please provide a description. Is there anything a future user could do to mitigate these undesirable harms?

\textcolor{blue}{\textbf{Answer: }} Not to the authors' knowledge.

\textbf{Are there tasks for which the dataset should not be used?}
% If so, please provide a description.

\textcolor{blue}{\textbf{Answer: }} No.

\textbf{Any other comments?}

\textcolor{blue}{\textbf{Answer: }} No.

\subsection{Distribution}

\textbf{Will the dataset be distributed to third parties outside of the entity (e.g., company, institution, organization) on behalf of which the dataset was created?}
% If so, please provide a description.

\textcolor{blue}{\textbf{Answer: }} Yes, the dataset will be freely available to everyone.

\textbf{How will the dataset will be distributed (e.g., tarball on website, API, GitHub)?}
% Does the dataset have a digital object identifier (DOI)?

\textcolor{blue}{\textbf{Answer: }} Tarball on website.

\textbf{When will the dataset be distributed?}

\textcolor{blue}{\textbf{Answer: }} The dataset is freely availble as of June 12, 2023.

\textbf{Will the dataset be distributed under a copyright or other intellectual property (IP) license, and/or under applicable terms of use (ToU)?}
% If so, please describe this license and/or ToU, and provide a link or other access point to, or otherwise reproduce, any relevant licensing terms or ToU, as well as any fees associated with these restrictions.

\textcolor{blue}{\textbf{Answer: }} The dataset is available under Creative Commons Attribution 4.0 International License.

\textbf{Have any third parties imposed IP-based or other restrictions on the data associated with the instances?}
% If so, please describe these restrictions, and provide a link or other access point to, or otherwise reproduce, any relevant licensing terms, as well as any fees associated with these restrictions.

\textcolor{blue}{\textbf{Answer: }} No.

\textbf{Do any export controls or other regulatory restrictions apply to the dataset or to individual instances?}
% If so, please describe these restrictions, and provide a link or other access point to, or otherwise reproduce, any supporting documentation.

\textcolor{blue}{\textbf{Answer: }} No.

\textbf{Any other comments?}

\textcolor{blue}{\textbf{Answer: }} No.

\subsection{Maintenance}
% These questions are intended to encourage dataset creators to plan for dataset maintenance and communicate this plan with dataset consumers.

\textbf{Who is supporting/hosting/maintaining the dataset?
How can the owner/curator/manager of the dataset be contacted (e.g., email address)?}

\textcolor{blue}{\textbf{Answer: }} The maintainer of the dataset is Frederik Warburg (frewar1905@gmail.com)

\textbf{Is there an erratum?}
% If so, please provide a link or other access point.

\textcolor{blue}{\textbf{Answer: }} No.

\textbf{Will the dataset be updated (e.g., to correct labeling errors, add new instances, delete instances)?}
% If so, please describe how often, by whom, and how updates will be communicated to users (e.g., mailing list, GitHub)?

\textcolor{blue}{\textbf{Answer: }} No.

\textbf{If the dataset relates to people, are there applicable limits on the retention of the data associated with the instances (e.g., were individuals in question told that their data would be retained for a fixed period of time and then deleted)?}
% If so, please describe these limits and explain how they will be enforced.

\textcolor{blue}{\textbf{Answer: }} No.

\textbf{Will older versions of the dataset continue to be supported/hosted/maintained?}
% If so, please describe how. If not, please describe how its obsolescence will be communicated to users.

\textcolor{blue}{\textbf{Answer: }} Yes.

\textbf{If others want to extend/augment/build on/contribute to the dataset, is there a mechanism for them to do so?}
% If so, please provide a description. Will these contributions be validated/verified? If so, please describe how. If not, why not? Is there a process for communicating/distributing these contributions to other users? If so, please provide a description.

\textcolor{blue}{\textbf{Answer: }} No, this will be resolved on a case-by-case basis, as the nature of the dataset requires data collection events for expansion.

\textbf{Any other comments?}

\textcolor{blue}{\textbf{Answer: }} No.

\end{document}